\documentclass[lettersize,journal]{IEEEtran}
\usepackage{amsmath,amsfonts}
\usepackage{algorithmic}
\usepackage{algorithm}
\usepackage{array}
\usepackage[caption=false,font=normalsize,labelfont=sf,textfont=sf]{subfig}
\usepackage{textcomp}
\usepackage{stfloats}
\usepackage{url}
\usepackage{verbatim}
\usepackage{graphicx}
\usepackage{cite}
\usepackage{makecell}
\usepackage{multirow}
\usepackage{booktabs}

\usepackage{adjustbox}
\usepackage{bbding}
\usepackage{algorithm}
\usepackage{algorithmic}
\usepackage{color,xcolor,colortbl}
\definecolor{lightgray}{gray}{0.95}
\definecolor{color3}{gray}{0.95}
\definecolor{rouse}{rgb}{0.981,0.961,0.941}
\newcommand{\etal}{\textit{et al}. }
\newcommand{\ie}{\textit{i}.\textit{e}.}
\newcommand{\eg}{\textit{e}.\textit{g}.}
\usepackage[pagebackref=true,breaklinks=true,colorlinks,bookmarks=false,citecolor=green ]{hyperref}
\newcommand{\widthscalefive}{0.25}
\newcommand{\widthscalefivereal}{0.5}

\begin{document}

\title{Meta-Learning based Degradation Representation for Blind Super-Resolution}

\author{Bin Xia, Yapeng Tian, Yulun Zhang, Yucheng Hang, \\Wenming Yang,~\IEEEmembership{Senior Member,~IEEE,} Qingmin Liao,~\IEEEmembership{Senior Member,~IEEE}
\thanks{This work was supported by the Natural Science Foundation of China(No.62171251 \& 62311530100), the Natural Science Foundation of Guangdong Province(No.2020A1515010711) and the Special Foundation for the Development of Strategic Emerging Industries of Shenzhen(JCYJ20200109143010272).}
\thanks{Bin Xia, Yucheng Hang, Wenming Yang, and Qingmin Liao are
with the Department of Electronic Engineering, Shenzhen International Graduate School, Tsinghua University, Shenzhen 518055, China
(e-mail: xiab20@mails.tsinghua.edu.cn; hangyc20@mails.tsinghua.edu.cn; 
yang.wenming@sz.tsinghua.edu.cn; liaoqm@tsinghua.edu.cn).
}

\thanks{Yapeng Tian is with the Department of Computer Science, The University of Texas at Dallas, Richardson, TX 75080, USA (e-mail: yapeng.tian@utdallas.edu).
}
\thanks{Y.~Zhang is with the Computer Vision Lab, ETH Z\"{u}rich, Z\"{u}rich 8092, Switzerland
(e-mail: yulun100@gmail.com).}
}

\markboth{Journal of \LaTeX\ Class Files,~Vol.~14, No.~8, August~2021}%
{Shell \MakeLowercase{\textit{et al.}}: A Sample Article Using IEEEtran.cls for IEEE Journals}


\maketitle

\begin{abstract}

Blind image super-resolution (blind SR)  aims to generate high-resolution (HR) images from low-resolution (LR) input images with unknown degradations. To enhance the performance of SR, the majority of blind SR methods introduce an explicit degradation estimator, which helps the SR model adjust to unknown degradation scenarios.  Unfortunately, it is impractical to provide concrete labels for the multiple combinations of degradations (\eg, blurring, noise, or JPEG compression) to guide the training of the degradation estimator. Moreover, the special designs for certain degradations hinder the models from being generalized for dealing with other degradations.
Thus, it is imperative to devise an implicit degradation estimator that can extract discriminative degradation representations for all types of degradations without requiring the supervision of degradation ground-truth. To this end, we propose a Meta-Learning based Region Degradation Aware SR Network (MRDA), including Meta-Learning Network (MLN), Degradation Extraction Network (DEN), and Region Degradation Aware SR Network (RDAN). To handle the lack of ground-truth degradation, we use the MLN to rapidly adapt to the specific complex degradation after several iterations and extract implicit degradation information. Subsequently, a teacher network MRDA$_{T}$ is designed to further utilize the degradation information extracted by MLN for SR. However, MLN requires iterating on paired LR and  HR images, which is unavailable in the inference phase. Therefore, we adopt knowledge distillation (KD) to make the student network learn to directly extract the same implicit degradation representation (IDR) as the teacher from LR images. Furthermore, we introduce an RDAN module that is capable of discerning regional degradations, allowing IDR to adaptively influence various texture patterns. Extensive experiments under classic and real-world degradation settings show that MRDA achieves SOTA performance and can generalize to various degradation processes.  
\end{abstract}


\vspace{-1mm}
\section{Introduction}
\label{sec:intro}

Single Image Super-Resolution (SISR), aiming to recover the realistic and detailed high-resolution (HR) counterpart from a low-resolution (LR) image, is a long-standing problem in low-level vision. Recently, CNN-based methods have achieved remarkable success on the SISR task for the powerful feature representation capability of deep neural networks~\cite{SRCNN,VDSR,EDSR,SRGAN,LapSRN,Memnet,ESRGAN,TIP1,TIP2,TIP3,TIP4}. Nevertheless, these methods will suffer a severe performance drop when the degradation model difference exists between the training and inference data~\cite{IKC}.

\begin{figure*}[t]
	\begin{center}
		\includegraphics[width=0.95\textwidth]{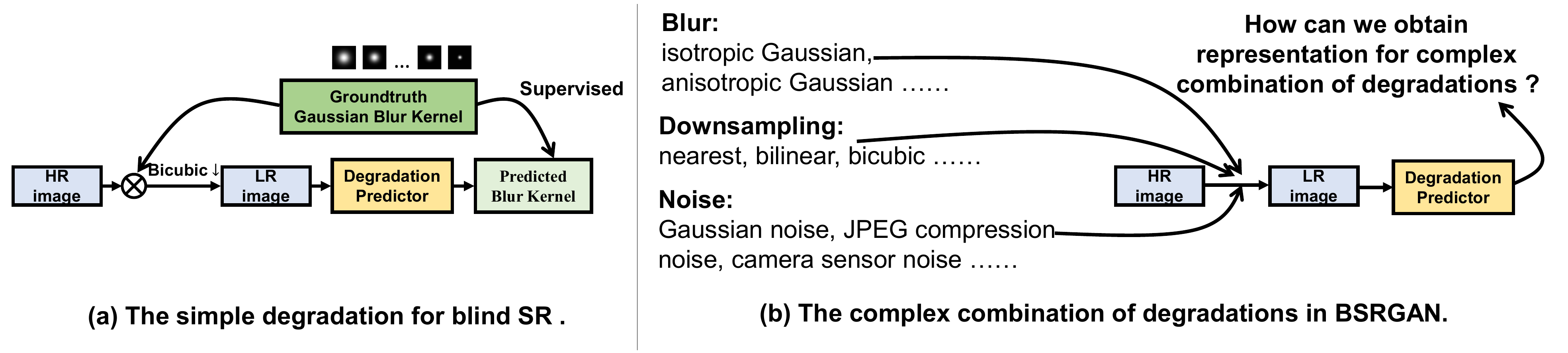}
	\end{center}
	\vspace{-5mm}
	\caption{(a) Some blind SR~\cite{IKC,DAN} methods used simple Gaussian blur kernel as degradation to obtain LR images from HR images. Then, they train the degradation predictor with the supervision of corresponding ground-truth blur kernels. (b) For the complex combination of multiple degradations in BSRGAN~\cite{BSRGAN} or Real-ESRGAN~\cite{Real-ESRGAN}, how can we obtain a concrete representation for supervising the training of the degradation estimator? }
	\label{fig:motivation}
\end{figure*} 

To address the problems, SR methods can be divided into three categories: \textbf{(1)} Several SR methods~\cite{SRMDNF,ZSSR,USRNet,MZSR,UDVD} directly take the known degradation information as prior. When the true degradation information is provided, these non-blind methods can achieve promising SR results. \textbf{(2)} Most classic blind SR methods~\cite{IKC,DAN,ZSSR,MZSR,yildirim2022iterative} assumes the classic image degradation process (Eq.~\eqref{equal:form}), and introduce explicit degradation estimator~\cite{nonparametric,kernelgan} to estimate blur degradation information for non-blind SR networks~\cite{SRMDNF,ZSSR,USRNet}. By introducing a degradation estimator, non-blind SR networks can adapt to specific degradation from huge degradation space. For instance, IKC~\cite{IKC} took Gaussian blur as degradation and adopted an iterative kernel correction strategy to estimate blur kernel information. \textbf{(3)} Recently, BSRGAN~\cite{BSRGAN} and Real-ESRGAN~\cite{Real-ESRGAN} found that classic image degradation process cannot cover real-world degradation, and classic blind SR methods~\cite{IKC,DAN} fail to produce visually pleasant results for most real images such as JPEG compressed ones. Thus, they develop a new and
complex degradation process to better cover real-world degradation space, which forms a variant of blind SR called real-world super-resolution (Real-SR). As shown in Fig.~\ref{fig:motivation} (b), BSRGAN~\cite{BSRGAN} and Real-ESRGAN~\cite{Real-ESRGAN} performed randomly shuffle on comprehensive degradations (such as blur, noise, down-sampling, and JPEG compression) in different severity to synthesize LR images covering the degradation space of real images. 

Although explicit degradation estimators in classic blind SR methods can help to boost SR performance, they have two restrictions: \textbf{(1)} It is challenging to provide precise ground-truth labels to represent the multiple degradation combination (\eg, degradations in BSRGAN and Real-ESRGAN) for supervising explicit degradation estimator training. \textbf{(2)} Explicit estimation errors may lead to SR failure.  Hence, it is necessary to explore implicit degradation representation (IDR) for blind SR. Recently, DASR~\cite{DASR} adopts contrastive learning~\cite{MOCO} realizing IDR learning. However, DASR does not have supervision and merely roughly distinguishes degradations by pushing away or pulling close features, which cannot fully capture discriminative enough degradation characteristics for blind SR. 

In this paper, we develop a novel Meta-Learning based Region Degradation Aware SR Network (MRDA) to implicitly extract degradation-specific information for restoration.  As shown in Fig.~\ref{fig:all_proccess}, our MRDA consists of Meta-Learning Network (MLN), Degradation Extraction Network (DEN), and Region Degradation Aware SR Network (RDAN). Our MRDA has three training stages: \textbf{(1)} When a network is trained on a specific degradation, it can perform well on this degradation since the network can extract the specific implicit degradation information. However, it is impossible to train networks for each specific degradation from scratch, which requires massive iterations and computational resources. To address the problem, we adopt meta-learning~\cite{MAML} on a lightweight network (MLN) to enforce the network to rapidly adapt to specific degradation with several iterations and extract corresponding implicit degradation representation (IDR).  \textbf{(2)} In training stage 2, we train 
a teacher network MRDA$_{T}$ with the IDR extracted by MLN. This makes MRDA$_{T}$ use IDR for blind SR.  \textbf{(3)} However, MLN needs to finetune on the paired ground-truth HR and LR image to adapt to specific degradation. However, the ground-truth HR image is unavailable in the inference phase. Hence, in training stage 3, we make the MRDA$_{S}$ learn to directly extract the same IDR as  MRDA$_{T}$ from LR images. In Sec.~\ref{sec:ablation}, we visualize the distribution of degradation representations with t-SNE~\cite{T-SNE}, and our model can learn discriminative IDR.

Moreover, we develop a Region Degradation Aware Block (RDA Block) to exploit the extracted degradation representation for restoration. The motivation is that the same degradation has different impacts on different texture patterns. For example, applying a blur kernel has little effect on flat regions but can significantly affect sharp textures. Experiments on both synthetic and real images show that the proposed MRDA achieves state-of-the-art blind SR performance. Our main contributions are threefold:  
\begin{itemize}
	\item We propose MRDA, a strong, simple, and efficient baseline for blind SR, that can generalize to any degradation process, which addresses the weakness of explicit degradation estimation. To the best of our knowledge, the design of IDR based blind SR networks has received little attention so far.
 
	\item We develop the RDA block to adaptively adjust the influence of IDR with spatial modulation coefficients, which can fully use IDR for blind SR. 
	
	\item  We validate MRDA on two different degradation settings: classic degradation and real-world SR. Our model achieves SOTA performance, indicating that MRDA is capable of handling various types of degradations.
	
\end{itemize}

\section{Related Work}
\label{sec:formatting}

\subsection{Non-blind Super-Resolution}
\label{sec:problem}
In the past few years, numerous efforts have been made to improve performance on restoring HR images from LR images synthesized by bicubic interpolation. As a pioneer work, SRCNN~\cite{SRCNN} utilizes a three-layer network to learn an end-to-end image mapping function between LR and HR images. Since then, CNN-based SR methods~\cite{VDSR,EDSR,SRGAN,LapSRN,ESRGAN,SAN} have achieved promising performance and have been widely studied. To further enhance the restoration of details, RCAN~\cite{RCAN} and NLRN~\cite{NLRN} adopt channel attention and non-local attention, respectively. Furthermore, VGG~\cite{vgg} based perceptual loss and GANs~\cite{WGAN-GP} based adversarial loss are applied in SR Network and show visually satisfying results. To compress the model size for deployment, \cite{distill1} and \cite{distill2} use knowledge distillation~\cite{hinton2015distilling} on the SR network. Besides, to improve the performance of SR, \cite{distill3} learns additional privileged information by exploiting knowledge distillation. 

However, when the degradation difference exists between training and testing images, the performance of the above SR methods is unsatisfactory~\cite{IKC}. Thus, several methods~\cite{SRMDNF,USRNet,UDVD} employ known degradation as an additional input and show great superiority on multiple-degradation SR. Moreover, ZSSR~\cite{ZSSR}  involves training during the inference phase using a degraded LR image as input. Consequently, ZSSR can adapt to the specific degradation existing in the LR image. Nevertheless, ZSSR requires thousands of iterations for convergence. Therefore, MZSR~\cite{MZSR} and MLSR~\cite{meta-fast} introduce the meta-learning scheme~\cite{meta1,meta2,meta3,MAML,MAML++} to make the SR network can adapt to a specific degradation within several iterations. 

\subsection{Blind Super-Resolution}

The aforementioned non-blind SR methods~\cite{SRMDNF,ZSSR} rely on the known degradation existing in given LR images.
To address unknown degradation, blind SR~\cite{liu2022blind} has emerged. Specifically, IKC~\cite{IKC} iteratively corrects the estimated degradation according to previously produced SR results. DAN~\cite{DAN} applied an alternating scheme to estimate degradation and generate SR results. Recently, 
BSRGAN~\cite{BSRGAN} and Real-ESRGAN~\cite{Real-ESRGAN} developed highly intricate and comprehensive degradation procedures that encompass the degradation space of real images. This has given rise to a new version of blind SR, known as real-world SR.

Previous blind SR methods~\cite{IKC, DAN} have typically relied on designing explicit degradation estimators for each specific degradation type and process. However, these explicit degradation estimators often suffer from complexity issues, making them difficult to generalize to other degradation types. Furthermore, it can be challenging to provide accurate ground-truth labels for multiple degradation combinations~\cite{Real-ESRGAN}.
Recently, Wang~\etal proposed an implicit degradation representation method~\cite{DASR} using contrastive learning to distinguish different degradations. However, for the lack of supervision, DASR is unstable and cannot fully extract discriminative degradation representation for guiding blind SR. Different from these methods using explicit degradation estimator or contrastive learning, we propose MRDA to extract  discriminative IDR with meta-learning.

\section{Methodology}

\begin{table}[t]
  \centering
  \caption{The implications of symbols in our paper.}
  \resizebox{1\linewidth}{!}{
    \begin{tabular}{l|l}
    \toprule[0.2em]
    Symbol &  Implication \\
    \midrule
    $I_{HR}$   & Ground-truth high-resolution images \\
    $I_{LR}$   & Low-resolution images \\
    $I_{SR}$   & Super-resolved images \\
    \multirow{2}[0]{*}{MRDA$_T$} & \multirow{2}[0]{*}{\shortstack{Teacher meta learning based region\\degradation aware SR network}} \\
          &  \\
    \multirow{2}[0]{*}{MRDA$_S$} & \multirow{2}[0]{*}{\shortstack{Student meta learning based region\\degradation aware SR network}} \\
          &  \\
    MLN   & Meta learning network \\
    DEN$_T$ & Teacher Degradation Extraction Network \\
    DEN$_S$ & Student Degradation Extraction Network \\
    RDAN  & Region Degradation aware SR network \\
    IDR   & Implicit degradation representation \\
    $\mathbf{D}_{T}^{\prime}$   &  IDR extracted by MLN \\
    $\mathbf{D}_{T}$    & Compressed IDR by DEN$_{T}$ \\
    $\mathbf{D}_{S}$    & Compressed IDR extracted by DEN$_{S}$ \\
    \midrule
    $I_{HR}^{tr}$ & HR images used for Meta Training \\
    $I_{HR}^{te}$ & HR images used for Meta Testing \\
    $I_{SR}^M$  & SR images generated by MLN \\
    \bottomrule[0.2em]
    \end{tabular}%
    }
    \label{tab:mysymbol}%
  
\end{table}%

\subsection{Overview}

For better clarity, we initially provide implications of symbols in Tab.~\ref{tab:mysymbol}.

Blind SR methods can be categorized into two groups: classic blind SR and real-world blind SR. We will validate our MRDA on both classic blind SR and real-world blind SR scenarios, in order to demonstrate its ability to generalize across different types of degradation and adapt to complex real-world degradation. 

\textbf{Classic blind SR}~\cite{DASR,IKC} typically employ the classic image degradation process, as defined by.
\begin{equation}
I_{LR}=\left(I_{HR} \otimes k\right) \downarrow_{s}+n,
\label{equal:form}
\end{equation}
where $\otimes$ indicates convolution operation, $k$ is blur kernel, $\downarrow_{s}$ refers to downsampling operation with scale factor $s$, and $n$ usually represents additive white Gaussian noise. 

Classic blind SR networks~\cite{IKC,DAN,DASR} are commonly trained with $\mathcal{L}_{rec}$.  
\begin{equation}
\label{eq:rec}
\mathcal{L}_{rec}=\left\|I_{HR}-I_{SR}\right\|_{1},
\end{equation}
where $I_{HR}$ and $I_{SR}$ are ground-truth HR and SR images respectively.

\textbf{Real-world SR} is a variant of classic blind SR that utilizes more intricate degradation procedures. 
The real-world blind SR approaches~\cite{Real-ESRGAN, BSRGAN} involve comprehensive degradation operations, including blur, noise, down-sampling, and JPEG compression. The severity of each operation is controlled by randomly sampling the corresponding hyper-parameters, the degradation orders are randomly shuffled, and second-order degradation is introduced to increase the complexity of the degradation. Since degradation is complex and cannot provide specific degradation labels, they directly use SR networks without degradation estimators. Their SR networks~\cite{Real-ESRGAN, BSRGAN} emphasize visual quality trained with $\mathcal{L}_{vis}$. 
\begin{equation}
\label{eq:vis}
\mathcal{L}_{vis}=\mathcal{L}_{rec}+\mathcal{L}_{per}+\mathcal{L}_{adv},
\end{equation}
where $\mathcal{L}_{per}$ and $\mathcal{L}_{adv}$ refer to the perceptual loss~\cite{perceptual} and adversarial loss~\cite{Real-ESRGAN}, respectively.

\begin{figure*}[t]
	\begin{center}
		\includegraphics[width=1\textwidth]{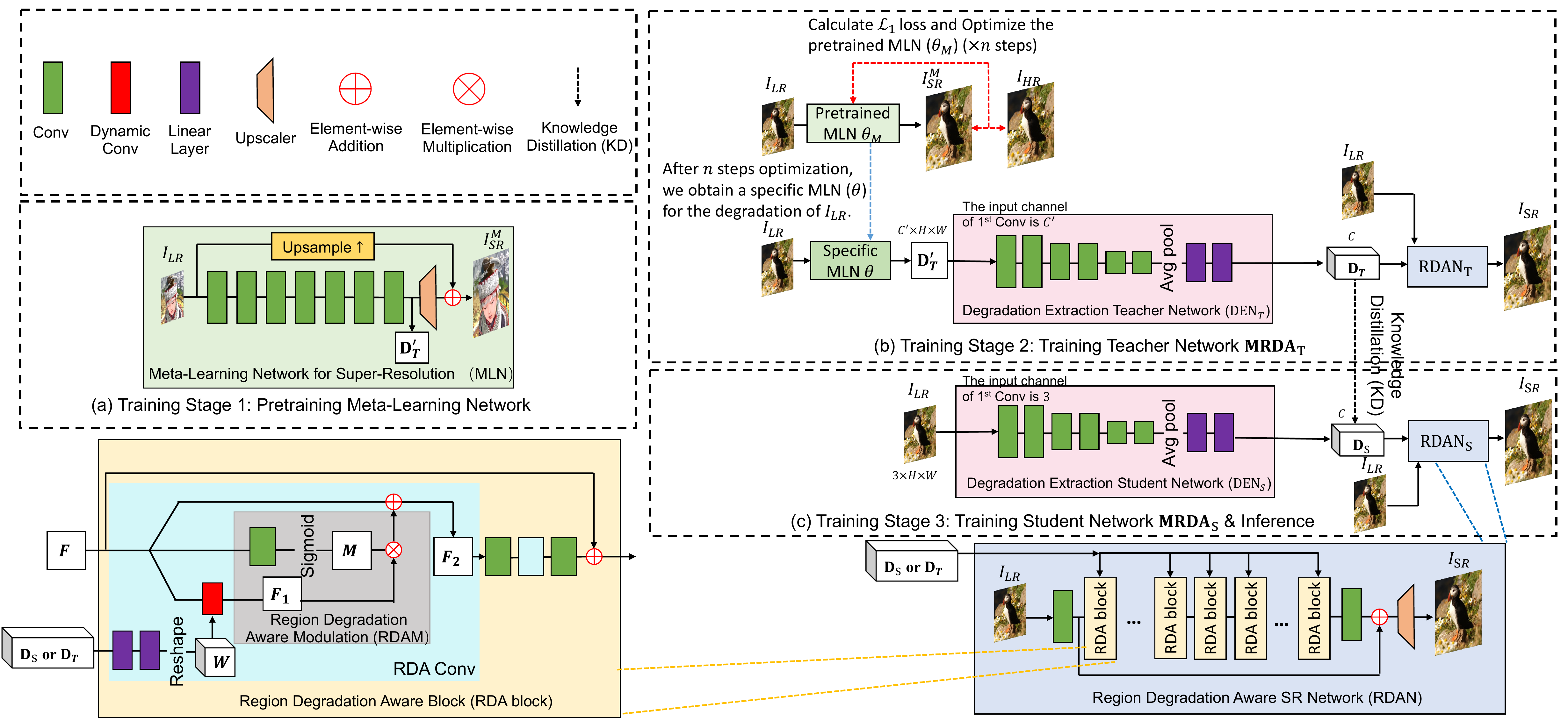}
	\end{center}
        \vspace{-3mm}
	\caption{The overview of our proposed Meta-Learning based Region Degradation Aware SR Network (MRDA). Our MRDA consists of Meta-Learning Network (MLN), Degradation Extraction Network (DEN), and Region Degradation Aware SR Network (RDAN). (a) We use meta-learning to pretrain MLN, which enables it to adapt to any degradation in just a few iterations. (b) Training MRDA$_{T}$. Given paired LR and ground-truth HR image, we obtain the specific MLN adapting to degradation of the LR image by optimizing pretrained MLN with several iterations. Then, we use the DEN$_{T}$ to compress IDR. The compressed IDR is sent into RDAN$_{T}$ to guide restoration. (c) We train DEN$_{S}$ to directly extract the same IDR as DEN$_{T}$ from LR images. It is notable that DEN$_{T}$ and DEN$_{S}$ share the same structure, except for the input channels of the first Conv layer. In DEN$_{T}$, the input channels of the first Conv layer are $C^{\prime}$, whereas in DEN$_{S}$, it is 3. Our MRDA uses LeakyReLU (negative slope is set to 0.1) as the non-linearity layer and does not adopt any normalization layer.}
	\label{fig:all_proccess}

\end{figure*}

Next, we will introduce the overall process of MRDA. As shown in Fig.~\ref{fig:all_proccess}, our MRDA consists of MLN, DEN, and RDAN.  As we can see, our training phase has three stages: \textbf{(1)} Training models from scratch for specific degradation requires massive iterations. To reduce computational costs, we pretrain MLN with meta-learning scheme MAML~\cite{MAML} to make MLN able to adapt to a specific degradation rapidly (Fig.~\ref{fig:all_proccess} (a) ). More details are given in Sec.~\ref{sec:meta}. \textbf{(2)} We train teacher network MRDA$_{T}$ by exploiting the intermediate feature map $\mathbf{D}^{\prime}_{T}$ of MLN as degradation information (Fig.~\ref{fig:all_proccess}~(b) ). Notably, in this stage, MLN only performs meta-learning inference. More details are given in Sec.~\ref{sec:teacher}. \textbf{(3)} MRDA$_{T}$ requires MLN to perform meta-learning inference by iterating on ground-truth HR images and LR images. However, the groud-truth HR images are unavailable in the inference stage. Therefore, as shown in Fig.~\ref{fig:all_proccess}~(c), we train student network MRDA$_{S}$ by initializing RDAN$_{S}$ with weights of the RDAN$_{T}$ and performing knowledge distillation between DEN$_{T}$ and DEN$_{S}$ to force DEN$_{S}$ to directly extract the same IDR $\mathbf{D}_{S}$ from LR image. More details are given in Sec.~\ref{sec:distillation}.  

In the inference phase, we only take the $I_{LR}$ as the input and only use MRDA$_{S}$ to perform blind SR (Fig.~\ref{fig:all_proccess}~(c)), which  is simple and efficient. Moreover, in Sec.~\ref{sec:RDAN}, we will introduce our RDAN in detail.   

\begin{algorithm}[t]
	\caption{Pretraining Meta-Learning Network}
	\label{alg:meta-train}
	\textbf{Input}: High-resolution dataset $\mathcal{D}_{\mathcal{HR}}$,  degradation process $p(D)$ (\ie, classic degradation or real-world degradation in Sec.~\ref{sec:data_prdeal}), the number of gradient updates $n$, the number of tasks $m$, and learning rates $\alpha$,$\beta$\\
	\textbf{Output}: MLN parameters $\theta_{M}$
 
	\begin{algorithmic}[1] 
		\STATE Randomly initialize MLN parameters with $\theta_{M}$ 
		\STATE Pretrain with bicubicly downsampling degradation
		\FOR{each epoch }
		\STATE Sample task batch $I_{HR}^{tr}$ and $I_{HR}^{te}$ from $\mathcal{D}_{\mathcal{HR}}$
            \STATE \textbf{Meta Training}
		\FOR{$i$ in $m$ tasks }
		\STATE Generate $I_{LR}^{tr}(i)$ and $I_{LR}^{te}(i)$ with $I_{HR}^{tr}(i)$ and $I_{HR}^{te}(i)$ under $p(D)$, respectively.
		\STATE $\theta_{i}\leftarrow \theta_{M}$.
		\FOR{$n$ steps }
		\STATE Compute adapted parameters \textbf{not in Upscaler} with gradient descent: $I_{SR}^{M}(i)=\operatorname{MLN}(I_{LR}^{tr}(i);\theta_{i})$, $\theta_{i}=\theta_{i}-\alpha \nabla_{\theta_{i}} \mathcal{L}_{1}(I_{SR}^{M},I_{HR}^{tr}(i))$
		\ENDFOR
		\ENDFOR
            \STATE \textbf{Meta Testing}
		\STATE Update parameters \textbf{not in Upscaler} with average test loss: $I_{SR}^{M}(i)=\operatorname{MLN}(I_{LR}^{te}(i);\theta_{i})$,
		$\theta_{M} \leftarrow \theta_{M}-\beta \nabla_{\theta_{M}} \sum_{i =[1, m]} \mathcal{L}_{1}(I_{SR}^{M}(i),I_{HR}^{te}(i))$
		\ENDFOR
	\end{algorithmic}
\end{algorithm}

\subsection{Pretraining Meta-Learning Network }
\label{sec:meta}
Before introducing pretraining meta-learning, we would like to emphasize our motivation.

The network trained and tested with a single degradation can extract accurate implicit degradation information and achieve promising results~\cite{xie2021finding}.
Thus, we can attempt to use the property to extract implicit degradation information. However,  there are infinite types of degradation, making it impractical to train a network from scratch for each individual type due to the massive computational costs involved. To mitigate the issue, as shown in Fig.~\ref{fig:all_proccess}~(a), we propose a lightweight MLN. MLN is pretrained with meta-learning~\cite {MAML} scheme to force the network able to adapt to specific degradation in several iterations. It is notable that, since degradation always has a significant impact on high-frequency parts of images, we adopt a global residual connection on MLN to filter out the most low-frequency information related to image content and make the network concentrate on high-frequency details for extracting degradation information  (Fig.~\ref{fig:all_proccess} (a) green box). The MLN consists of 8 convolutional layers and an upscaler, which is able to adapt to specific degradation with negligible computations.

The pretraining process of MLN is illustrated in Alg.~\ref{alg:meta-train} and Fig.~\ref{fig:all_proccess}~(a). Specifically, we first pretrain MLN on DIV2K~\cite{DIV2K} and Flickr2K~\cite{Flickr2K} with known ``bicubic" degradation and optimize it by minimizing the pixel-wise  $\mathcal{L}_{1}$ loss between predicted SR images $I_{SR}^{M}$ and ground-truth $I_{HR}$ (Alg.~\ref{alg:meta-train} L2), which can expedite the subsequent meta-learning training on multiple degradations. Then, we randomly select $m$ degradation types from degradation process $p(D)$ (\ie, classic degradation or real-world degradation in Sec.~\ref{sec:data_prdeal}) and sample $I_{HR}^{tr}$ and $I_{HR}^{te}$ for meta training and testing (Alg.~\ref{alg:meta-train} L7). For each degradation type ($i\in [1,m]$), we generate corresponding training LR images $I_{LR}^{tr}$ and $I_{LR}^{te}$ and assign the MLN with parameters $\theta_{M}$ (Alg.~\ref{alg:meta-train} L7, L8). After that, in meta-training, MLN iterates $n$ times on the paired $I_{HR}^{tr}$ and $I_{LR}^{tr}$ to obtain parameters $\theta_{i}$ for specific degradation (Alg.~\ref{alg:meta-train} L9-L12). Afterward, in meta-testing, we update the $\theta_{M}$ according to each degradation-specific $\theta_{i}$ as MAML \cite{MAML} did (Alg.~\ref{alg:meta-train} 14). Note that MLN fixes parameters of Upscaler after ``bicubic" degradation pretraining to keep the extracted implicit degradation information $\mathbf{D}^{\prime}_{T} $ in the same domain.

\subsection{Training Teacher Network MRDA$_{T}$ }
\label{sec:teacher}

\begin{algorithm}[t]
	\caption{Training teacher network MRDA$_{T}$ }
	\label{alg:MRDA}
	\textbf{Input}: High-resolution dataset $\mathcal{D}_{\mathcal{HR}}$, degradation process $p(D)$ (\ie, classic degradation or real-world degradation in Sec.~\ref{sec:data_prdeal}), Meta-Learning Network parameters trained $\theta_{M}$, the number of gradient updates $n$, and learning rates $\alpha$, $\gamma$\\
	\textbf{Output}: MRDA$_{T}$ parameters $\theta_{T}$ 
 
	\begin{algorithmic}[1] 
		\STATE Randomly initialize MRDA$_{T}$ parameters with $\theta_{T}$
		\FOR{each epoch}
		\STATE Generate batch $I_{HR}$ and $I_{LR}$ with $\mathcal{D}_{\mathcal{HR}}$ and $p(D)$ \\
		\STATE Initialize Meta-Learning Network: $\theta\leftarrow \theta_{M}$ \\
		\FOR{$n$ steps}
	    \STATE $I_{SR}^{M}=\operatorname{MLN}\left(I_{LR};\theta\right)$
		\STATE Evaluate loss $\mathcal{L}(\theta)=\left\|I_{HR}-I_{SR}^{M}\right\|_{1}$ \\
		\STATE Compute adapted parameters \textbf{not in Upscaler} with gradient descent: $\theta=\theta-\alpha \nabla_{\theta} \mathcal{L}(\theta)$
		\ENDFOR
		\STATE Obtain specific $\mathbf{D}^{\prime}_{T} $ from $\operatorname{MLN}\left(I_{LR};\theta\right)$ \\
		\STATE Compute adapted parameters with ADAM~\cite{adam}: 
		
		$\theta_{T} \leftarrow \theta_{T}-\gamma \nabla_{\theta_{T}} \mathcal{L}\left(\operatorname{MRDA}_{T}(I_{LR},\mathbf{D}^{\prime}_{T};\theta_{T}), I_{HR}\right)$, where $\mathcal{L}$ can refer to either $\mathcal{L}_{rec}$ ( Eq.~\eqref{eq:rec}) for classic degradation, or to $\mathcal{L}_{vis}$ ( Eq.~\eqref{eq:vis}) for real-world degradation.
		\ENDFOR
	\end{algorithmic}
\end{algorithm}

As shown in Fig.~\ref{fig:all_proccess}~(b), we train the MRDA$_{T}$ to fully exploit the IDR extracted by MLN for restoration. Specifically, given paired LR and ground-truth HR images, the pretrained MLN can rapidly adapt to the degradation of LR images by iterating several steps. After obtaining the specific MLN (adapting to the degradation of given LR images), we use this specific MLN to extract the corresponding IDR $\mathbf{D}^{\prime}_{T}\in \mathbb{R}^{ C^{\prime} \times H \times W}$ from given LR images. After that, the DEN$_{T}$ further compress the IDR $\mathbf{D}^{\prime}_{T}$ to  $\mathbf{D}_{T}\in \mathbb{R}^{ C}$ for RDAN$_{T}$ to restore LR images. It is notable that the reason we compress the IDR  into a simplified vector form is to reduce the computational cost of converting it into dynamic convolutional weights of RDAN.

The detailed training process of training MRDA$_{T}$ is shown in Alg.~\ref{alg:MRDA} and Fig.~\ref{fig:all_proccess}~(b). We first randomly select degradations and ground-truth HR images from $p(D)$ (\ie, classic degradation or real-world degradation in Sec.~\ref{sec:data_prdeal}) and $\mathcal{D}_{\mathcal{HR}}$ separately and generate corresponding $I_{LR}$ (Alg.~\ref{alg:MRDA} L3). Given paired LR and ground-truth HR images, the pretrained MLN can rapidly adapt to the degradation of LR images by $n$  iterations and obtain specific MLN (Alg.~\ref{alg:MRDA} L5-L9). After that, we use specific MLN to extract the IDR $\mathbf{D}^{\prime}_{T}\in \mathbb{R}^{ C^{\prime} \times H \times W}$ from given LR images (Alg.~\ref{alg:MRDA} L10).  Afterward, we input $\mathbf{D}^{\prime}_{T}$ and $I_{LR}$ to MRDA$_{T}$, take $I_{HR}$ as ground-truth, and adopt $\mathcal{L}$ loss as the loss function to train MRDA$_{T}$ (Alg.~\ref{alg:MRDA} L11), where $\mathcal{L}$ can represent either $\mathcal{L}_{rec}$ ( Eq.~\eqref{eq:rec}) for classic blind SR or  $\mathcal{L}_{vis}$ ( Eq.~\eqref{eq:vis}) for real-world SR.

\subsection{Training Student Network MRDA$_{S}$ }
\label{sec:distillation}

The extraction of the IDR in MRDA$_{T}$ relies on a specific MLN, which is obtained through an iterative process using LR and ground-truth HR images. However, during inference, the ground-truth HR is unknown, making it challenging to utilize the specific MLN. To address this issue, we introduce the student network DEN$_{S}$, which is trained to directly extract the same IDR as the specific MLN and DEN$_{T}$ from LR images. We accomplish this by employing the concept of knowledge distillation (KD) in our network. Different from traditional KD tasks~\cite{hinton2015distilling, distill1, distill5, distill6, distill7} that typically focus on model compression, our KD aims to enable DEN$_{S}$ to learn how to extract the IDR from LR images as accurately as the specific MLN and DEN$_{T}$.

The training process of MRDA$_{S}$ is shown in Fig.~\ref{fig:all_proccess}~(c).  Specifically, we initialize the parameters of the student network RDAN$_{S}$ with those of the teacher network RDAN$_{T}$. This transfers the reconstruction capability of the teacher to the student and provides a good starting point for optimization. Then, we use the DEN$_S$ to extract $\mathbf{D}_{S}\in \mathbb{R}^{ C}$. Furthermore, we utilize the Kullback Leibler divergence ($\mathcal{L}_{kl}$) and  $\mathcal{L}_{abs}$ loss as KD loss to enforce the $\mathbf{D}_{S}$ extracted by DEN$_{S}$ close to $\mathbf{D}_{T}$ in distribution and absolute value separately. The $\mathcal{L}_{kl}$ and $\mathcal{L}_{abs}$ can be formulated as Eq.~\eqref{eq:kl} and Eq.~\eqref{eq:abs}, respectively.

\begin{equation}
\label{eq:kl}
\begin{aligned}
\mathcal{L}_{kl} &= \frac{1}{4C}\sum_{i=1}^{4C} \mathbf{D}_{Tnorm}(i) \log \left(\frac{\mathbf{D}_{Tnorm}(i)}{\mathbf{D}_{Snorm}(i)}\right),
\end{aligned}
\end{equation}
\begin{equation}
\label{eq:abs}
\begin{aligned}
\mathcal{L}_{abs} &= \frac{1}{4C}\sum_{i=1}^{4C}\left\|\mathbf{D}_{T}(i)-\mathbf{D}_{S}(i)\right\|_{1},
\end{aligned}
\end{equation}
where $\mathbf{D}_{T}$ and $\mathbf{D}_{S}\in \mathbb{R}^{ C}$ are both intermediate IDR. $\mathbf{D}_{Tnorm}$ and $\mathbf{D}_{Snorm}\in \mathbb{R}^{ C}$ are normalized with softmax operation. 

For the \textbf{classic blind SR}, following previous classic blind SR works~\cite{IKC,DASR}, we add KD losses to $\mathcal{L}_{rec}$ (Eq.~\eqref{eq:rec}).
The resulting overall loss function of MRDA$_{S}$ as $\mathcal{L}_{classic}$:
\begin{equation}
\label{eq:all}
\mathcal{L}_{classic}=\mathcal{L}_{rec}+\lambda_{kl}\mathcal{L}_{kl}+\lambda_{abs}\mathcal{L}_{abs},
\end{equation}
where $\lambda_{kl}$ and $\lambda_{abs}$ will balance the importance of the two distillation losses.

For the \textbf{real-world SR}, we further add KD losses to $\mathcal{L}_{vis}$ (Eq.~\eqref{eq:vis}). The  overall loss function of MRDA$_{S}$ as $\mathcal{L}_{real}$:
\begin{equation}
\label{eq:real}
\mathcal{L}_{real}=\mathcal{L}_{vis}+\lambda_{kl}\mathcal{L}_{kl}+\lambda_{abs}\mathcal{L}_{abs}.
\end{equation}

\begin{table*}[t]
	\centering
	\caption{4$\times$ SR quantitative comparison on datasets with Gaussian8~\cite{IKC} kernels. The bottom two methods marked in rouse use implicit representation degradation to guide blind SR. The Mult-Adds and runtime are computed based on an LR size of $180\times320$. }
	\resizebox{1\linewidth}{!}{
	\begin{tabular}{lccccccccccccc}
		\toprule[0.2em]
		\multirow{2}[2]{*}{Methods} & \multirow{2}[2]{*}{Param (M)} & \multirow{2}[2]{*}{Mult-Adds (G)} & \multirow{2}[2]{*}{Time (ms)} & \multicolumn{2}{c}{Set5~\cite{Set5}} & \multicolumn{2}{c}{Set14~\cite{Set14}} & \multicolumn{2}{c}{BSD100~\cite{B100}} & \multicolumn{2}{c}{Urban100~\cite{Urban100}} & \multicolumn{2}{c}{Manga109~\cite{Manga109}} \\
		&       &       &       & PSNR  & SSIM  & PSNR  & SSIM  & PSNR  & SSIM  & PSNR  & SSIM  & PSNR  & SSIM \\
		\midrule
		\midrule
		Bicubic & -     & -     & -     & 24.57 & 0.7108 & 22.79 & 0.6032 & 23.29 & 0.5786 & 20.35 & 0.5532 & 21.50  & 0.6933 \\
		\midrule
		RCAN~\cite{RCAN}  & 15.59 & 1082.41 & 546.04 & 26.60 & 0.7598 & 24.85 & 0.6513 & 25.01 & 0.6170 & 22.19 & 0.6078 & 23.52 & 0.7428 \\
		\midrule
		SwinIR~\cite{swinir}  & 11.90 & - & 328.15 & 26.63	& 0.7606 &	24.86 &	0.6524 &	25.02 &	0.6182 &	22.22	& 0.6093 &	23.55 &	0.7448  \\
		\midrule
		Bicubic+ZSSR~\cite{ZSSR} & 0.23  & -     & 30946.60 & 26.45 & 0.7279 & 24.78 & 0.6268 & 24.97 & 0.5989 & 21.11 & 0.5805 & 23.53 & 0.724 \\
		\midrule
		IKC~\cite{IKC}   & 5.32  & 2528.03 & 1043.62 & 31.67 & 0.8829 & 28.31 & 0.7643 & 27.37 & 0.7192 & 25.33 & 0.7504 & 28.91 & 0.8782 \\
		\midrule
		DAN~\cite{DAN} & 4.33  & 1098.33 & 201.04 & 31.89 & 0.8864 & 28.42 & 0.7687 & 27.51 & 0.7248 & 25.86 & 0.7721 & 30.50  & 0.9037 \\
		\midrule
		AdaTarget~\cite{AdaTarget} & 16.7  & 1032.59 & 109.77 & 31.58 & 0.8814 & 28.14 & 0.7626 & 27.43 & 0.7216 & 25.72 & 0.7683 & 29.97 & 0.8955 \\
		\midrule
		\midrule
		\rowcolor{rouse}
		DASR~\cite{DASR} & 5.84  & 185.66 & 44.14 & 31.46 & 0.8789 & 28.11 & 0.7603 & 27.44 & 0.7214 & 25.36  & 0.7506 & 29.39 & 0.8861 \\
		\midrule
		\rowcolor{rouse}
		MRDA$_{S}$ (Ours) & 5.84  & 172.13 & 42.61 & 31.98 & 0.8872 & 28.42 & 0.7671 & 27.55 & 0.7254 & 25.90  & 0.7734 & 30.51 & 0.9088 \\
		\bottomrule[0.2em]
	\end{tabular}%
}
\vspace{-2mm}
	\label{tab:iso_show}%
\end{table*}%

\begin{figure*} [t]
        \newlength\fsdttwofigBD
	\setlength{\fsdttwofigBD}{-1.5mm}
	\centering
	\resizebox{1\linewidth}{!}{
		\begin{tabular}{cccccccc}
			
			\includegraphics[width=0.192\textwidth]{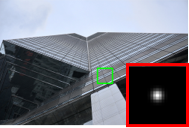} &
			\includegraphics[width=0.13\textwidth]{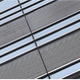} & \includegraphics[width=0.13\textwidth]{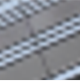} &
			\includegraphics[width=0.13\textwidth]{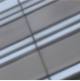} &
			\includegraphics[width=0.13\textwidth]{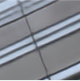} &
			\includegraphics[width=0.13\textwidth]{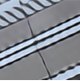} &
			\includegraphics[width=0.13\textwidth]{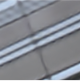} &
			\includegraphics[width=0.13\textwidth]{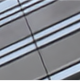} \\

			 & HR & Bicubic & RCAN~\cite{RCAN} & IKC~\cite{IKC} & DAN~\cite{DAN} & DASR~\cite{DASR} & Ours	\\

			\includegraphics[width=0.192\textwidth]{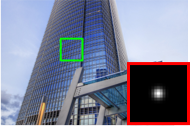} &
			\includegraphics[width=0.13\textwidth]{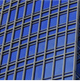} & \includegraphics[width=0.13\textwidth]{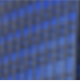} &
			\includegraphics[width=0.13\textwidth]{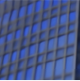} &
			\includegraphics[width=0.13\textwidth]{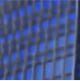} &
			\includegraphics[width=0.13\textwidth]{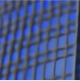} &
			\includegraphics[width=0.13\textwidth]{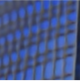} &
			\includegraphics[width=0.13\textwidth]{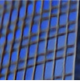} \\
			
			 & HR & Bicubic & RCAN~\cite{RCAN} & IKC~\cite{IKC} & DAN~\cite{DAN} & DASR~\cite{DASR} & Ours	\\
			 
			 \includegraphics[width=0.192\textwidth]{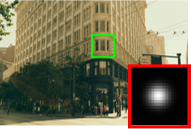} &
			 \includegraphics[width=0.13\textwidth]{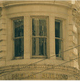} & \includegraphics[width=0.13\textwidth]{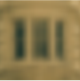} &
			 \includegraphics[width=0.13\textwidth]{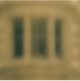} &
			 \includegraphics[width=0.13\textwidth]{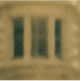} &
			 \includegraphics[width=0.13\textwidth]{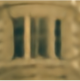} &
			 \includegraphics[width=0.13\textwidth]{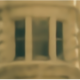} &
			 \includegraphics[width=0.13\textwidth]{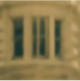} \\
			 
			 & HR & Bicubic & RCAN~\cite{RCAN} & IKC~\cite{IKC} & DAN~\cite{DAN} & DASR~\cite{DASR} & Ours	\\
			 \includegraphics[width=0.192\textwidth]{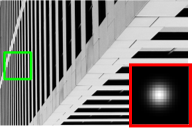} &
				\includegraphics[width=0.13\textwidth]{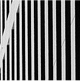} & \includegraphics[width=0.13\textwidth]{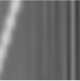} &
				\includegraphics[width=0.13\textwidth]{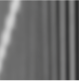} &
				\includegraphics[width=0.13\textwidth]{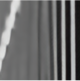} &
				\includegraphics[width=0.13\textwidth]{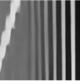} &
				\includegraphics[width=0.13\textwidth]{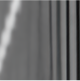} &
				\includegraphics[width=0.13\textwidth]{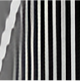} \\
				
			 & HR & Bicubic & RCAN~\cite{RCAN} & IKC~\cite{IKC} & DAN~\cite{DAN} & DASR~\cite{DASR} & Ours	\\
			 
			 \includegraphics[width=0.192\textwidth]{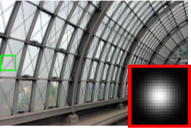} &
			 \includegraphics[width=0.13\textwidth]{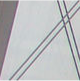} & \includegraphics[width=0.13\textwidth]{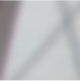} &
			 \includegraphics[width=0.13\textwidth]{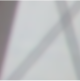} &
			 \includegraphics[width=0.13\textwidth]{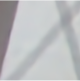} &
			 \includegraphics[width=0.13\textwidth]{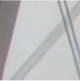} &
			 \includegraphics[width=0.13\textwidth]{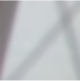} &
			 \includegraphics[width=0.13\textwidth]{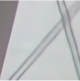} \\
			 
			 & HR & Bicubic & RCAN~\cite{RCAN} & IKC~\cite{IKC} & DAN~\cite{DAN} & DASR~\cite{DASR} & Ours	\\
			 
			 \includegraphics[width=0.192\textwidth]{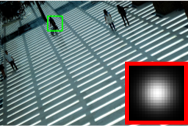} &
			 \includegraphics[width=0.13\textwidth]{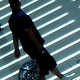} & \includegraphics[width=0.13\textwidth]{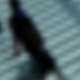} &
			 \includegraphics[width=0.13\textwidth]{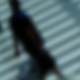} &
			 \includegraphics[width=0.13\textwidth]{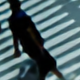} &
			 \includegraphics[width=0.13\textwidth]{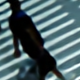} &
			 \includegraphics[width=0.13\textwidth]{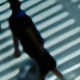} &
			 \includegraphics[width=0.13\textwidth]{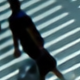} \\
			 
			 & HR & Bicubic & RCAN~\cite{RCAN} & IKC~\cite{IKC} & DAN~\cite{DAN} & DASR~\cite{DASR} & Ours	\\

		\end{tabular}}
	\caption{Visual comparison achieved on Urban100  with Gaussian8~\cite{IKC} kernels for 4$\times$ SR. The blur kernels are displayed with red boxes.}
	\label{fig:iso_show}
\end{figure*}

\subsection{Region Degradation Aware SR Network}
\label{sec:RDAN}
After introducing three training stages, we will proceed to introduce the structure of our Region Degradation Aware SR Network (RDAN). Since different texture patterns have different responses for the same degradation, we propose RDAN to generate modulation coefficients that can fully use extracted IDR for SR recovery.  As shown in Fig.~\ref{fig:all_proccess} blue box,  RDAN consists of 27 RDA blocks.

Within each RDA block, as shown in Fig.~\ref{fig:all_proccess} yellow box, two RDA convolutional layers are employed to aggregate implicit degradation representation (IDR). Motivated by \cite{DASR}, one branch of RDA convolutional layers adopts dynamic convolution, which learns to predict the weights of convolution kernels conditioned on the IDR. RDA block uses the IDR $\mathbf{D}\in \mathbb{R}^{ C}$ extracted by DEN$_{T}$ or DEN$_{S}$ from LR image to guide SR recovery.  Specifically, RDA Conv uses $\mathbf{D}_{T}$ or $\mathbf{D}_{S}$ to generate weights for degradation-specific convolution weights $\mathbf{w}$, and applies dynamic convolution on input features $\mathbf{F}\in \mathbb{R}^{ C \times H \times W}$ as Eq.~\eqref{eq:d1} to obtain $\mathbf{F}_{1}\in \mathbb{R}^{ C \times H \times W}$. Another branch of RDA convolutional layers uses the residual connection. To adapt to different texture patterns, the Region Degradation Aware Modulation (RDAM) module of our RDA block learns to generate modulation coefficient maps $\mathbf{M}\in \mathbb{R}^{ C \times H \times W}$ based on the $\mathbf{F}_{1}$ as Eq.~\eqref{eq:d2}. These modulation coefficient maps are applied to the dynamic convolution branch to adjust the influence of IDR as Eq.~\eqref{eq:d3}. 
\begin{equation}
\label{eq:d1}
\mathbf{F}_{1} = \operatorname{DynamicConv}(\mathbf{F};\mathbf{w}),
\end{equation}
\begin{equation}
\label{eq:d2}
\mathbf{M} = \operatorname{Sigmoid}(\operatorname{Conv}(\mathbf{F})),
\end{equation}
\begin{equation}
\label{eq:d3}
\mathbf{F}_{2} = \mathbf{M}\cdot \mathbf{F}_{1} +\mathbf{F},
\end{equation}
where $\mathbf{F}, \mathbf{F}_{1}, \mathbf{F}_{2}\in \mathbb{R}^{ C \times H \times W}$ are the input feature maps, intermediate feature maps, and output of RDA Conv, respectively. $\operatorname{DynamicConv}$ indicates the dynamic convolution, where we use the form of depthwise separable convolution~\cite{mobilenets} to further reduce computational costs. $\mathbf{w}\in \mathbb{R}^{ C \times 1 \times K_{h} \times K_{w}}$ is the depthwise separable dynamic convolution weights.  $\operatorname{Conv}$ represents  the convolution.

\section{Experiments}
\subsection{Data Preparation and Network Training}
\label{sec:data_prdeal}

We train and test our method on classic and real-world degradation settings. For the \textbf{classic degradation}, following previous works~\cite{DASR, IKC}, we combine 800 images in DIV2K~\cite{DIV2K} and 2,650 training images in Flickr2K~\cite{Flickr2K} as the training set. The training batch sizes are set to 64, and the LR patch sizes are 64$\times$64. The blur kernel size in Eq.~\eqref{equal:form} is fixed to 21$\times$21 following~\cite{IKC}.   \textbf{(1)} In Sec.~\ref{sec:V1}, we train and test on noise-free degradation with isotropic Gaussian kernels only. The kernel width $\sigma$ ranges are set to [0.2, 2.0] and [0.2, 4.0] for scale factors 2 and 4, respectively. We uniformly sample the kernel width in the above ranges. \textbf{(2)} In Sec.~\ref{sec:V2}, we train and test on degradation with anisotropic Gaussian kernels and noises. We use the additive Gaussian noise with covariance $\sigma=25$ and adopt anisotropic Gaussian kernels characterized by Gaussian probability density function $N(0, \Sigma)$ with zero mean and varying covariance matrix $\Sigma$. The covariance matrix $\Sigma$ is determined by two random eigenvalues $\lambda_{1}, \lambda_{2} \sim U(0.2,4)$ and a random rotation angle $\theta \sim U(0, \pi)$. In three training stages, we train all networks with 500 epochs. The initial learning rate $\beta$ for MLN, $\gamma$ for MRDA$_{T}$ and MRDA$_{S}$ training are all set to $2\times10^{-4}$ and decrease to half after every 200 epochs while the learning rate $\alpha$ for task-level learning is set to $10^{-2}$. For training MRDA$_{S}$, the  loss coefficient of  $\lambda_{kl}$ and $\lambda_{abs}$ are 1 and 0.01 separately for classic blind SR (Eq.~\eqref{eq:all}).

For the \textbf{real world degradation}, in Sec.~\ref{sec:V3}, similar to Real-ESRGAN~\cite{Real-ESRGAN}, we adopt DF2K and OutdoorSceneTraining~\cite{wang2018recovering} datasets for training. We train our network with the data synthesized by the complex combination of multiple degradation generation schemes in Real-ESRGAN~\cite{Real-ESRGAN}. In three training stages, we train MLN, MRDA$_{T}$, and MRDA$_{S}$ with 100K, 600K, and 600K iterations, respectively. The initial learning rate $\beta$ for MLN, $\gamma$ for MRDA$_{T}$ and MRDA$_{S}$ training are all set to $2\times10^{-4}$ and decrease to half after every $400K$ iterations while the learning rate $\alpha$ for task-level learning is set to $2\times10^{-2}$. For training MRDA$_{S}$, we set the  loss coefficient of  $\lambda_{kl}$ and $\lambda_{abs}$ to 1 and 0.01 respectively for real-world SR (Eq.~\eqref{eq:real}).

Our training process has 3 steps:  pertaining MLN, training MRDA$_{T}$, and training MRDA$_{S}$. During training, we randomly crop 64$\times$64 LR patches and their corresponding HR patches from the training examples and form a mini-batch of 64 images. The training images are further augmented via horizontal and vertical flipping and random rotation. We optimize the model by ADAM optimizer~\cite{adam} with $\beta_{1}=0.9$, $\beta_{2}=0.999$. In pretraining MLN, the number of gradient updates $n$ and tasks $m$ (Alg.~\ref{alg:meta-train}) in MLN and MRDA$_{T}$ learning are both $5$.

\begin{table*}[t]
	\centering
	\caption{PSNR results achieved on Set14~\cite{Set14} under anisotropic Gaussian blur and noises. The bottom two methods marked in rouse use implicit representation degradation to guide blind SR. The best results are marked in bold. The runtime is measured on an LR size of $180\times320$.}
	\resizebox{1\linewidth}{!}{
		\begin{tabular}{lcccccccccccc}
			\toprule[0.2em]
			\multicolumn{1}{c}{\multirow{3}[2]{*}{Method}} & \multicolumn{1}{c}{\multirow{3}[2]{*}{Params}} & \multicolumn{1}{c}{\multirow{3}[2]{*}{Time}} & \multicolumn{1}{c}{\multirow{3}[2]{*}{Noise}} & \multicolumn{9}{c}{Blur Kernel} \\
			&       &       &       & \multirow{2}[1]{*}{\includegraphics[height=0.6cm,width=0.6cm]{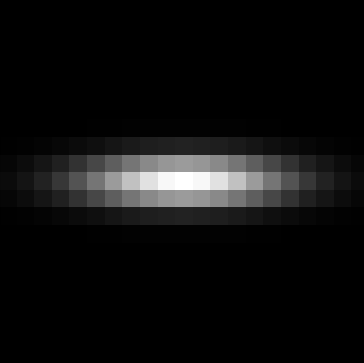}} & \multirow{2}[1]{*}{\includegraphics[height=0.6cm,width=0.6cm]{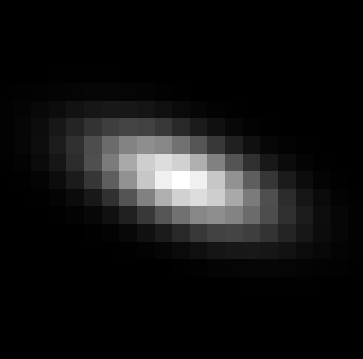}} & \multirow{2}[1]{*}{\includegraphics[height=0.6cm,width=0.6cm]{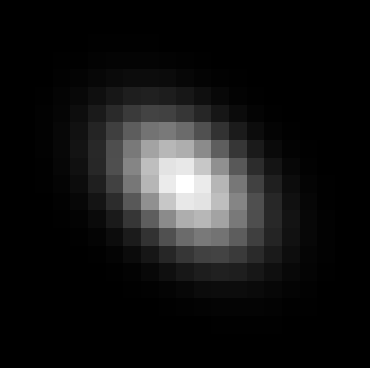}} & \multirow{2}[1]{*}{\includegraphics[height=0.6cm,width=0.6cm]{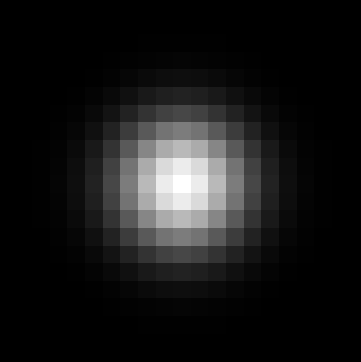}} & \multirow{2}[1]{*}{\includegraphics[height=0.6cm,width=0.6cm]{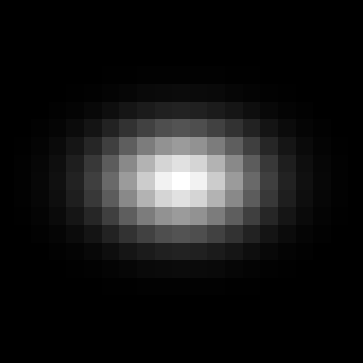}} & \multirow{2}[1]{*}{\includegraphics[height=0.6cm,width=0.6cm]{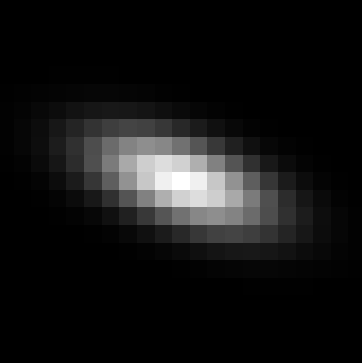}} & \multirow{2}[1]{*}{\includegraphics[height=0.6cm,width=0.6cm]{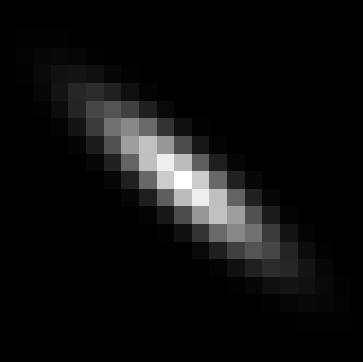}} & \multirow{2}[1]{*}{\includegraphics[height=0.6cm,width=0.6cm]{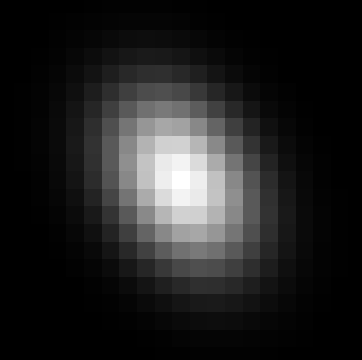}} & \multirow{2}[1]{*}{\includegraphics[height=0.6cm,width=0.6cm]{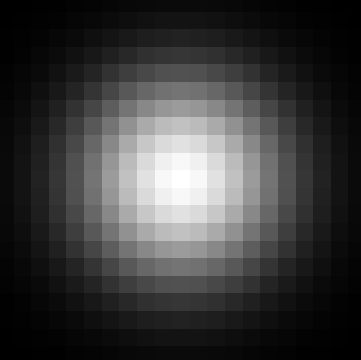}} \\
			&       &       &       &       &       &       &       &       &       &       &       &  \\
			\midrule
			\midrule
			\multicolumn{1}{c}{\multirow{3}[2]{*}{DnCNN~\cite{DnCNN} + RCAN~\cite{RCAN}}} & \multicolumn{1}{c}{\multirow{3}[2]{*}{650K+15.59M}} & \multicolumn{1}{c}{\multirow{3}[2]{*}{556.21ms}} & 0     & 24.28 &	24.47 &	24.6 &	24.64 &	24.58 &	24.47 &	24.31 &	23.97 &	23.01
			 \\
			&       &       & 10     & 23.88 &	24.03 &	24.16 &	24.21 &	24.13 &	24.03 &	23.88 &	23.62 &	22.76
			 \\
			&       &       & 20    & 23.45	& 23.58	& 23.70	& 23.73 &	23.69	& 23.57 & 	23.42	& 23.23	& 22.46
			 \\
			\midrule
			\multicolumn{1}{c}{\multirow{3}[2]{*}{DnCNN~\cite{DnCNN} + DAN~\cite{DAN}}} & \multicolumn{1}{c}{\multirow{3}[2]{*}{650K+4.33M}} & \multicolumn{1}{c}{\multirow{3}[2]{*}{211.21ms}} & 0     & 24.67	& 25.16 &	25.72 &	26.32 &	25.84 &	25.16 &	24.57 &	25.05 &	23.84
			 \\
			&       &       & 10     & 23.99 &	24.22 &	24.37 &	24.44 &	24.34 &	24.22 &	24.04 &	23.79 &	22.86
			 \\
			&       &       & 20    & 23.54	& 23.71 &	23.85 &	23.89 &	23.85 &	23.69 &	23.52 &	23.37 &	22.55
			 \\
			\midrule
			\multicolumn{1}{c}{\multirow{3}[2]{*}{DnCNN~\cite{DnCNN} +IKC~\cite{IKC}}} & \multicolumn{1}{c}{\multirow{3}[2]{*}{650K+5.32M}} & \multicolumn{1}{c}{\multirow{3}[2]{*}{1053.79ms}} & 0     & 24.76 &	25.55 &	26.54 &	27.33 &	26.55 &	25.55 &	24.64 &	25.99 &	25.49
			 \\
			&       &       & 10     & 24.20 &	24.54	&24.86 &	24.96&	24.78 &	24.52 &	24.23 &	24.19 &	23.14
			 \\
			&       &       & 20    & 23.62	& 23.87	& 24.07 &	24.15&	24.06 &	23.86 &	23.65 &	23.59 &	22.71
			 \\
			\midrule
			\midrule
			\multicolumn{1}{c}{\multirow{3}[2]{*}{DASR~\cite{DASR}}} & \multicolumn{1}{c}{\multirow{3}[2]{*}{5.84M}} & \multicolumn{1}{c}{\multirow{3}[2]{*}{44.14ms}} & \cellcolor{rouse}0     & \cellcolor{rouse}27.20 &	\cellcolor{rouse}27.62	&\cellcolor{rouse}27.74	&\cellcolor{rouse}27.85	&\cellcolor{rouse}27.82	&\cellcolor{rouse}27.62	&\cellcolor{rouse}27.38	&\cellcolor{rouse}27.44&	\cellcolor{rouse}26.27
			 \\
			&       &       & \cellcolor{rouse}10     & \cellcolor{rouse}25.26 &	\cellcolor{rouse}25.57 &	\cellcolor{rouse}25.64 &	\cellcolor{rouse}25.69 &	\cellcolor{rouse}25.62 &	\cellcolor{rouse}25.54 &	\cellcolor{rouse}25.42 &	\cellcolor{rouse}25.20 &	\cellcolor{rouse}24.37
			 \\
			&       &       &\cellcolor{rouse} 20    & \cellcolor{rouse}23.68	& \cellcolor{rouse}23.87 &	\cellcolor{rouse}24.20 &\cellcolor{rouse}	24.32 &	\cellcolor{rouse}24.09 &\cellcolor{rouse}	23.91 &	\cellcolor{rouse}23.76 &\cellcolor{rouse}	23.81 &	\cellcolor{rouse}22.87
			 \\
			\midrule
			
			\multicolumn{1}{c}{\multirow{3}[2]{*}{MRDA$_{S}$ (Ours)}} & \multicolumn{1}{c}{\multirow{3}[2]{*}{5.84M}} & \multicolumn{1}{c}{\multirow{3}[2]{*}{42.61ms}} & \cellcolor{rouse}0     & \cellcolor{rouse} \textbf{27.45} & \cellcolor{rouse} \textbf{27.85}  & \cellcolor{rouse}\textbf{28.04} & \cellcolor{rouse}\textbf{28.09} & \cellcolor{rouse}\textbf{28.01} & \cellcolor{rouse}\textbf{27.85} & \cellcolor{rouse}\textbf{27.64} & \cellcolor{rouse}\textbf{27.63} & \cellcolor{rouse}\textbf{26.29} \\
			&       &       & \cellcolor{rouse}10     & \cellcolor{rouse}\textbf{25.66} & \cellcolor{rouse}\textbf{25.81} & \cellcolor{rouse}\textbf{25.90} & \cellcolor{rouse}\textbf{25.92} & \cellcolor{rouse}\textbf{25.86} & \cellcolor{rouse}\textbf{25.80} & \cellcolor{rouse}\textbf{25.68} & \cellcolor{rouse}\textbf{25.41} & \cellcolor{rouse}\textbf{24.56} \\
			&       &       & \cellcolor{rouse}20    & \cellcolor{rouse}\textbf{24.71} & \cellcolor{rouse}\textbf{24.79} & \cellcolor{rouse}\textbf{24.87} & \cellcolor{rouse}\textbf{24.86} & \cellcolor{rouse}\textbf{24.90} &\cellcolor{rouse}\textbf{24.80} & \cellcolor{rouse}\textbf{24.69} & \cellcolor{rouse}\textbf{24.54} & \cellcolor{rouse}\textbf{23.81}  \\
			\bottomrule[0.2em]
		\end{tabular}%
	}
	\label{tab:aniso_show}%
\end{table*}%

\begin{figure*} [tb]
	\setlength{\fsdttwofigBD}{-1.5mm}
	\centering
	\resizebox{1\linewidth}{!}{
		\begin{tabular}{ccc}
			\includegraphics[width=0.25\textwidth]{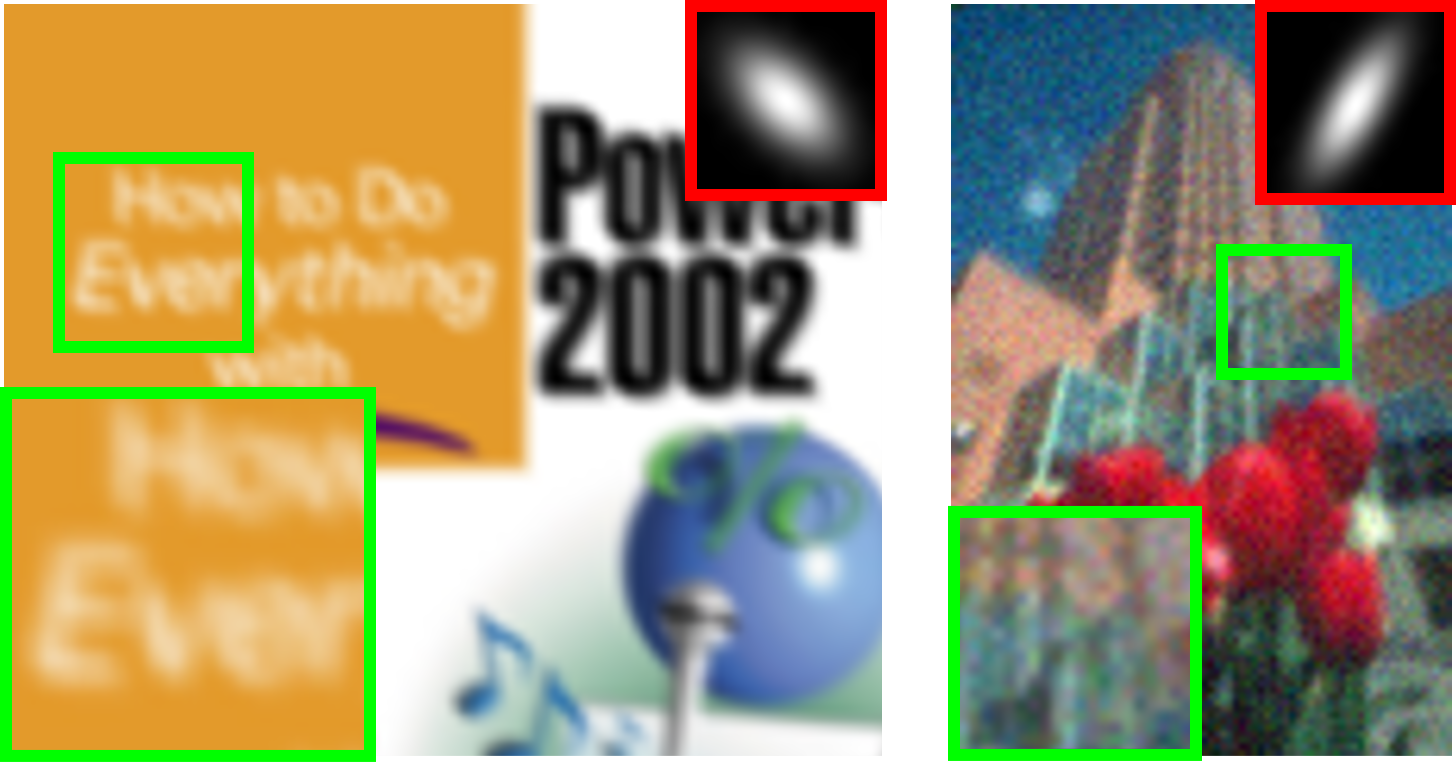} &
			\includegraphics[width=0.25\textwidth]{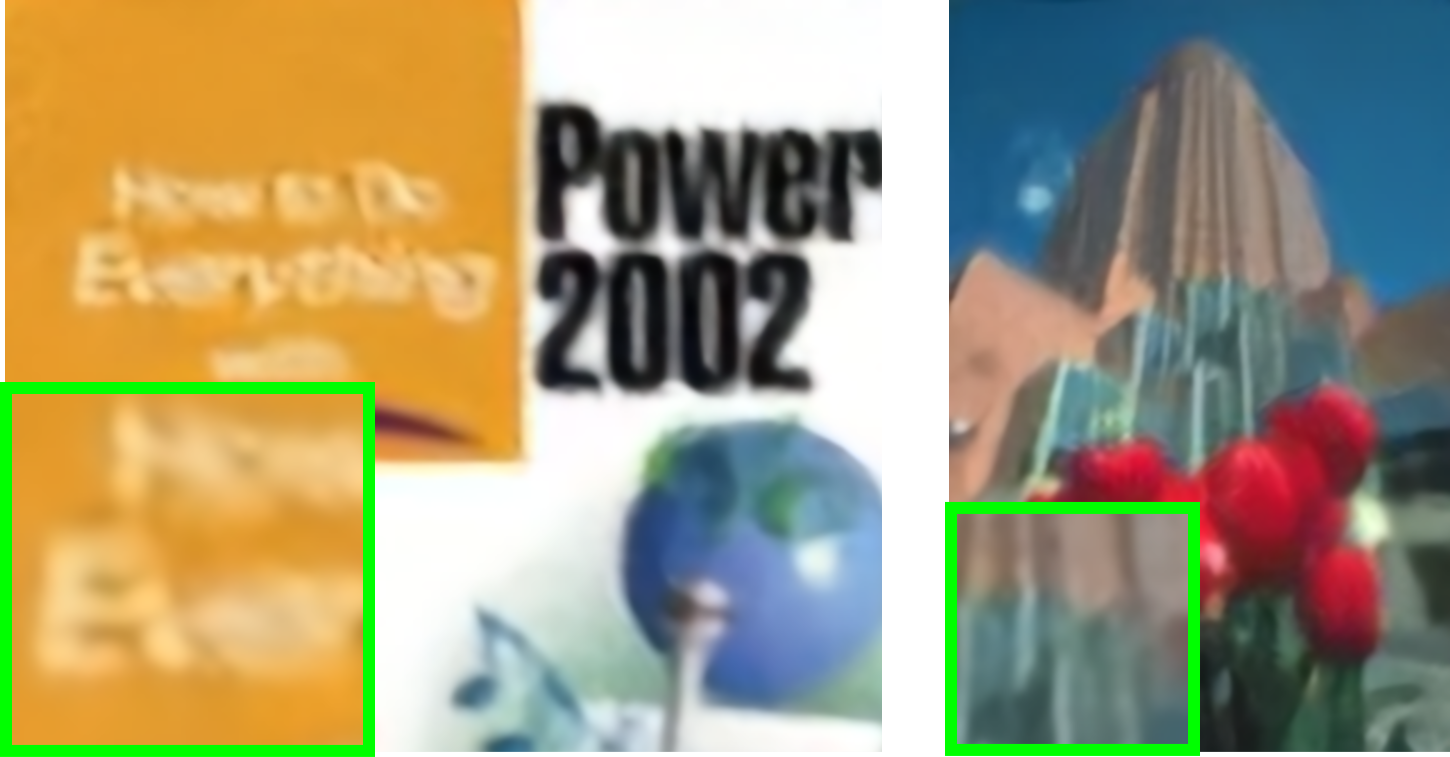} & \includegraphics[width=0.25\textwidth]{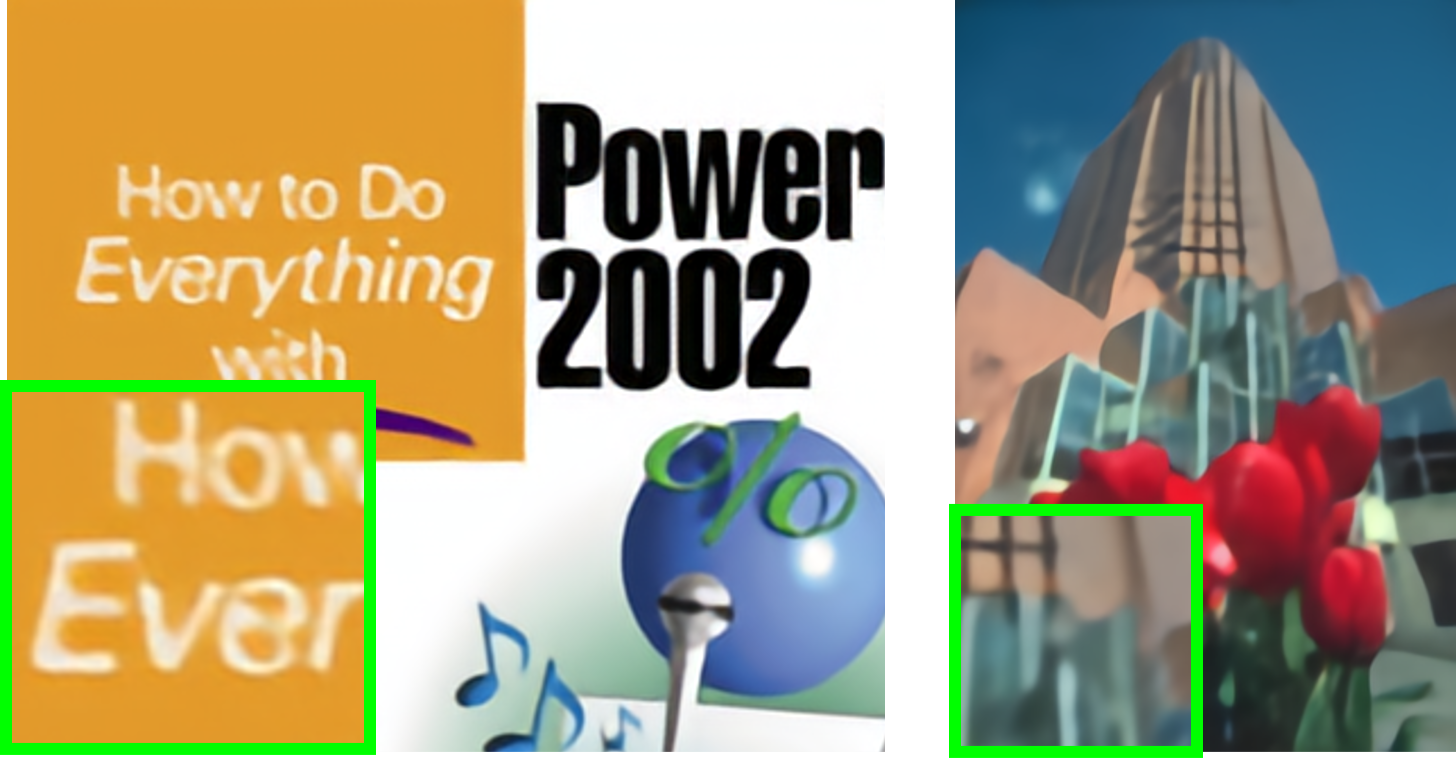}  \\			
			Bicubic & DnCNN~\cite{DnCNN} + RCAN~\cite{RCAN} & DASR~\cite{DASR}	\\
			\includegraphics[width=0.25\textwidth]{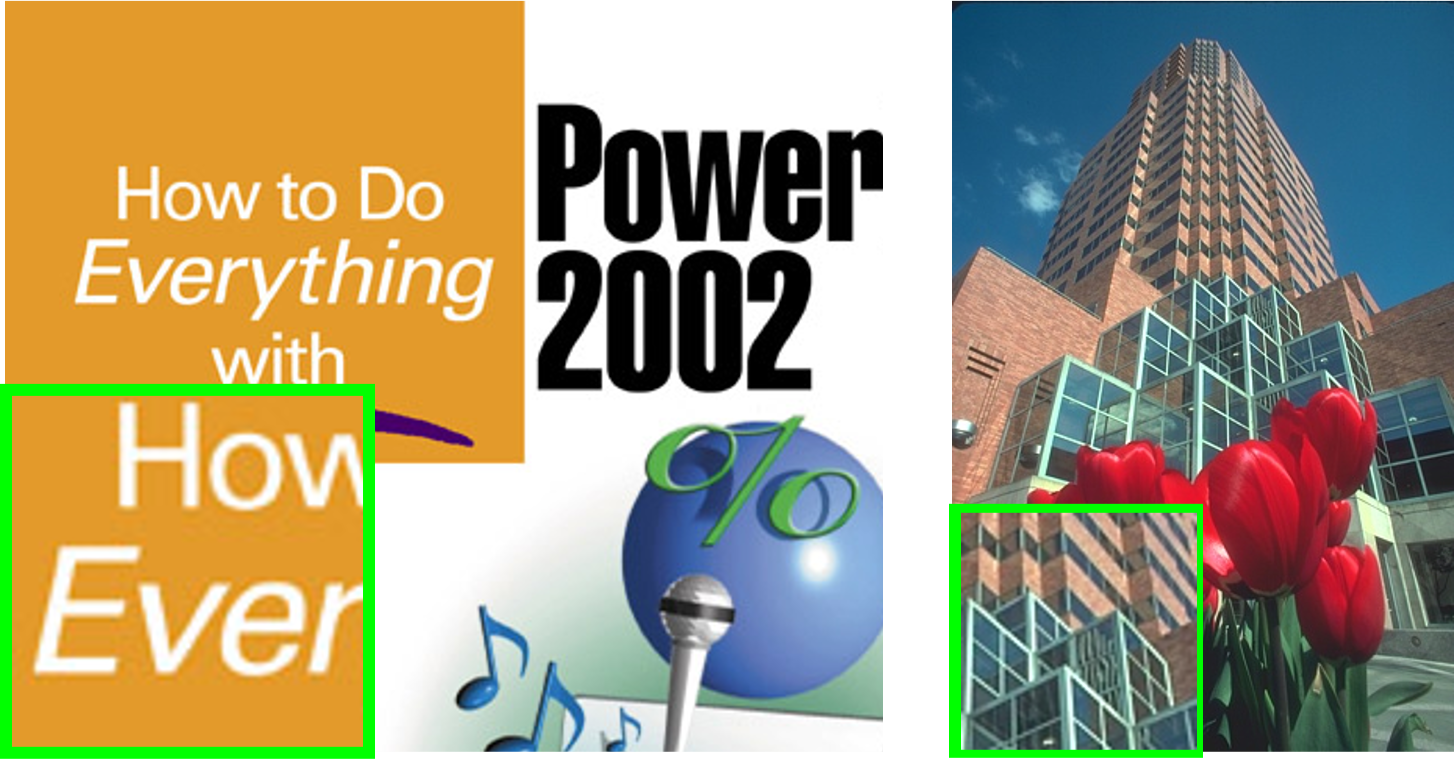} &
			\includegraphics[width=0.25\textwidth]{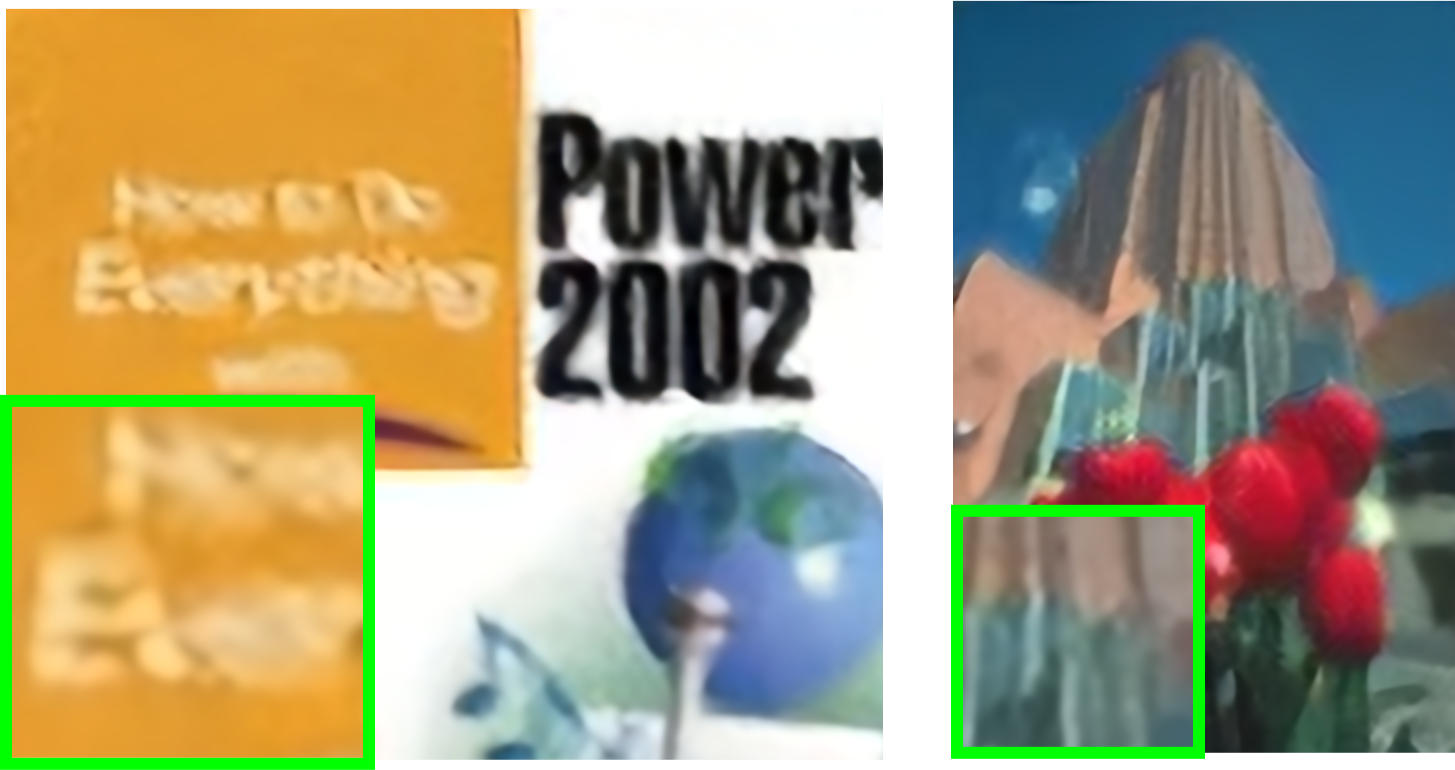} & \includegraphics[width=0.25\textwidth]{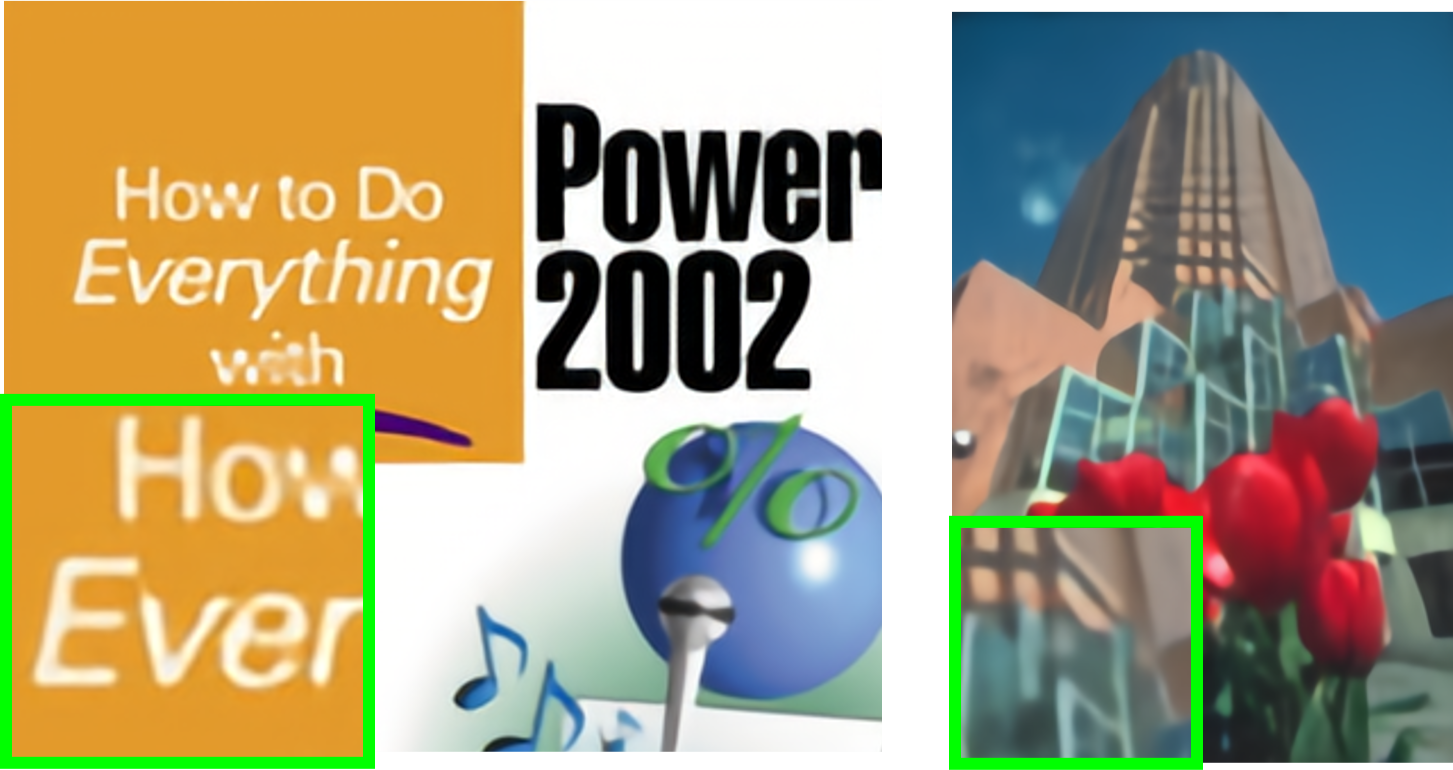}  \\			
			HR & DnCNN~\cite{DnCNN} + IKC~\cite{IKC} & Ours	\\
	\end{tabular}}
	\caption{Visual comparison for 4$\times$ SR on Set14 and BSD100. Noise levels are set to 0 and 10 for these two images separately. }
	\label{fig:aniso_show}
 \vspace{-2mm}
\end{figure*}

\begin{table*}[t]
  \centering
  \caption{4$\times$ SR quantitative comparison on real-world SR competition benchmarks. The Mult-Adds and runtime are computed based on an
LR size of 180 $\times$ 320. The best and second best performances are marked in bold and underlined. }
  \resizebox{1\linewidth}{!}{
    \begin{tabular}{l|ccc|ccc|ccc}
    \toprule[0.2em]
    \multirow{2}[2]{*}{Methods} & \multirow{2}[2]{*}{Parms (M)} & \multirow{2}[2]{*}{Mult-Adds(G)} & \multirow{2}[2]{*}{Runtime (ms)} & \multicolumn{3}{c|}{RealSRSet~\cite{RealSR}} & \multicolumn{3}{c}{NTIRE2020 Track1~\cite{NTIRE2020}} \\
          &       &       &       & LPIPS$\downarrow$ & PSNR$\uparrow$  & SSIM$\uparrow$  & LPIPS$\downarrow$ & PSNR$\uparrow$  & SSIM$\uparrow$ \\
    \midrule
    RealSR~\cite{RealSR} & 16.69 & 871.25 & 114.22 & 0.4428 & 24.72 & 0.7056 & 0.3648 & 26.50 & 0.7220 \\
    BSRGAN~\cite{BSRGAN} & 16.69 & 871.25 & 114.22 & 0.3648 & \underline{26.90}  & \underline{0.7912} & 0.3691 & \underline{26.75} & 0.7386 \\
    Real-ESRGAN~\cite{Real-ESRGAN} & 16.69 & 871.25 & 114.22 & 0.3629 & 26.07 & 0.7864 & 0.3471 & 26.40  & \underline{0.7431} \\
    MM-RealSR~\cite{MM-RealSR} & 26.13 & 1102.86 & 123.06 & \underline{0.3606} & 24.07 & 0.7708 & \underline{0.3446} & 25.19 & 0.7404 \\
    MRDA$_{s}$-GAN (Ours) & 19.59 & 572.96 & 63.41 & \textbf{0.3578} & \textbf{27.16}  & \textbf{0.8010} & \textbf{0.3313} & \textbf{27.04} & \textbf{0.7533} \\
    \bottomrule[0.2em]
    \end{tabular}%
    }
  \label{tab:real}%
\end{table*}%

\begin{figure*}[t]
    \newlength\fsdurthree
    \setlength{\fsdurthree}{0mm}
    \huge
    \centering
    \resizebox{1\linewidth}{!}{
        \begin{tabular}{cc}
            \begin{adjustbox}{valign=t}
                \begin{tabular}{c}
                \includegraphics[height=0.55\textwidth]{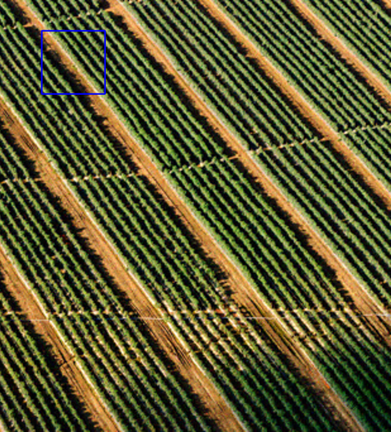}
                \end{tabular}
            \end{adjustbox}
            
            \begin{adjustbox}{valign=t}
                \begin{tabular}{ccc}
                    \includegraphics[width=\widthscalefive \textwidth]{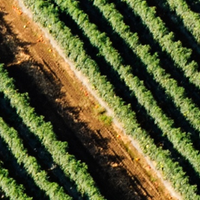} \hspace{\fsdurthree} &
                    \includegraphics[width=\widthscalefive \textwidth]{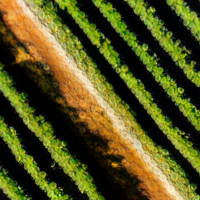} \hspace{\fsdurthree} &
                    \includegraphics[width=\widthscalefive \textwidth]{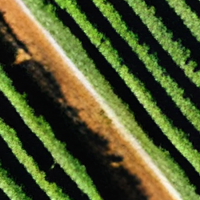} \hspace{\fsdurthree} 
                    \\
                    HR \hspace{\fsdurthree} &
                    \makecell{BSRGAN~\cite{BSRGAN}} \hspace{\fsdurthree} &
                    \makecell{MM-RealSR~\cite{MM-RealSR}} \hspace{\fsdurthree} 
                    \\
                    \includegraphics[width=\widthscalefive \textwidth]{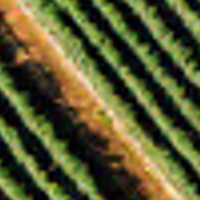} 
                    \hspace{\fsdurthree} &
                    \includegraphics[width=\widthscalefive \textwidth]{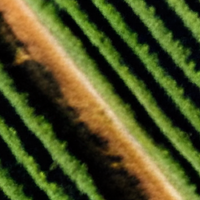} \hspace{\fsdurthree} &
                    \includegraphics[width=\widthscalefive \textwidth]{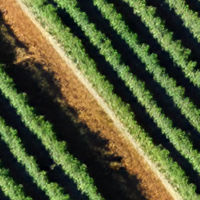} \hspace{\fsdurthree}
                    \\ 
                    LR \hspace{\fsdurthree} &
                    Real-ESRGAN~\cite{Real-ESRGAN} \hspace{\fsdurthree} &
                    \makecell{\textbf{MRDA$_{S}$-GAN}} \hspace{\fsdurthree} 
                \end{tabular}
            \end{adjustbox}

            \begin{adjustbox}{valign=t}
                \begin{tabular}{c}
                \includegraphics[height=0.55\textwidth]{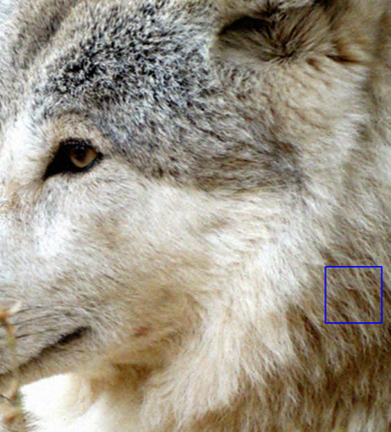}
                \end{tabular}
            \end{adjustbox}
            
            \begin{adjustbox}{valign=t}
                \begin{tabular}{ccc}
                    \includegraphics[width=\widthscalefive \textwidth]{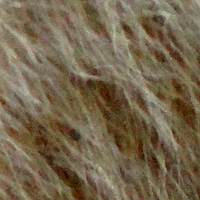} \hspace{\fsdurthree} &
                    \includegraphics[width=\widthscalefive \textwidth]{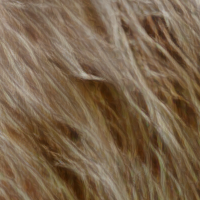} \hspace{\fsdurthree} &
                    \includegraphics[width=\widthscalefive \textwidth]{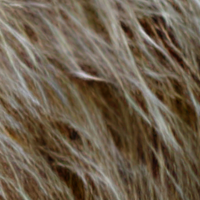} \hspace{\fsdurthree} 
                    \\
                    HR \hspace{\fsdurthree} &
                    \makecell{BSRGAN~\cite{BSRGAN}} \hspace{\fsdurthree} &
                    \makecell{MM-RealSR~\cite{MM-RealSR}} \hspace{\fsdurthree} 
                    \\
                    \includegraphics[width=\widthscalefive \textwidth]{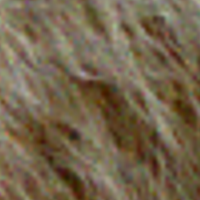} 
                    \hspace{\fsdurthree} &
                    \includegraphics[width=\widthscalefive \textwidth]{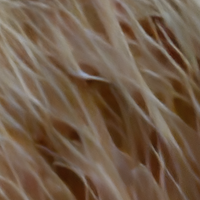} \hspace{\fsdurthree} &
                    \includegraphics[width=\widthscalefive \textwidth]{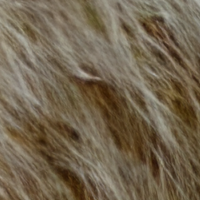} \hspace{\fsdurthree}
                    \\ 
                    LR \hspace{\fsdurthree} &
                    Real-ESRGAN~\cite{Real-ESRGAN} \hspace{\fsdurthree} &
                    \makecell{\textbf{MRDA$_{S}$-GAN}} \hspace{\fsdurthree} 
                \end{tabular}
            \end{adjustbox}
            \\
            \begin{adjustbox}{valign=t}
                \begin{tabular}{c}
                \includegraphics[height=0.55\textwidth]{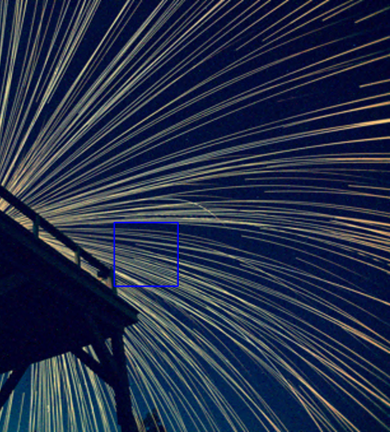}
                \end{tabular}
            \end{adjustbox}
            
            \begin{adjustbox}{valign=t}
                \begin{tabular}{ccc}
                    \includegraphics[width=\widthscalefive \textwidth]{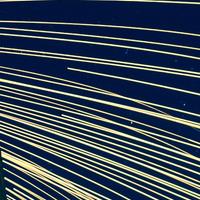} \hspace{\fsdurthree} &
                    \includegraphics[width=\widthscalefive \textwidth]{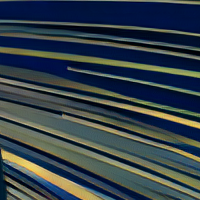} \hspace{\fsdurthree} &
                    \includegraphics[width=\widthscalefive \textwidth]{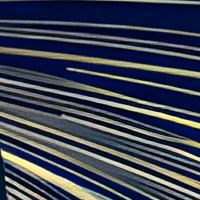} \hspace{\fsdurthree} 
                    \\
                    HR \hspace{\fsdurthree} &
                    \makecell{BSRGAN~\cite{BSRGAN}} \hspace{\fsdurthree} &
                    \makecell{MM-RealSR~\cite{MM-RealSR}} \hspace{\fsdurthree} 
                    \\
                    \includegraphics[width=\widthscalefive \textwidth]{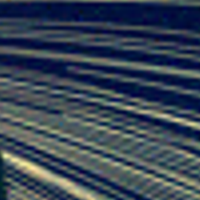} 
                    \hspace{\fsdurthree} &
                    \includegraphics[width=\widthscalefive \textwidth]{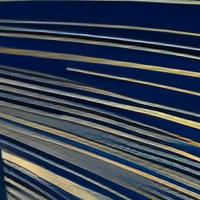} \hspace{\fsdurthree} &
                    \includegraphics[width=\widthscalefive \textwidth]{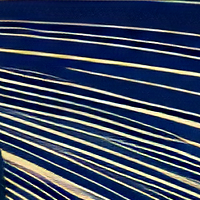} \hspace{\fsdurthree}
                    \\ 
                    LR \hspace{\fsdurthree} &
                    Real-ESRGAN~\cite{Real-ESRGAN} \hspace{\fsdurthree} &
                    \makecell{\textbf{MRDA$_{S}$-GAN}} \hspace{\fsdurthree} 
                \end{tabular}
            \end{adjustbox}

            \begin{adjustbox}{valign=t}
                \begin{tabular}{c}
                \includegraphics[height=0.55\textwidth]{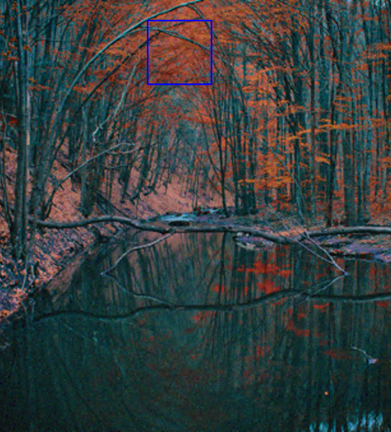}
                \end{tabular}
            \end{adjustbox}
            
            \begin{adjustbox}{valign=t}
                \begin{tabular}{ccc}
                    \includegraphics[width=\widthscalefive \textwidth]{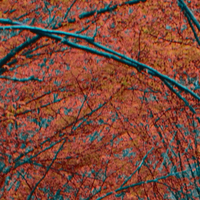} \hspace{\fsdurthree} &
                    \includegraphics[width=\widthscalefive \textwidth]{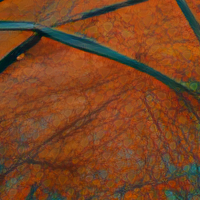} \hspace{\fsdurthree} &
                    \includegraphics[width=\widthscalefive \textwidth]{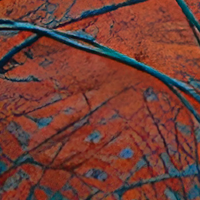} \hspace{\fsdurthree} 
                    \\
                    HR \hspace{\fsdurthree} &
                    \makecell{BSRGAN~\cite{BSRGAN}} \hspace{\fsdurthree} &
                    \makecell{MM-RealSR~\cite{MM-RealSR}} \hspace{\fsdurthree} 
                    \\
                    \includegraphics[width=\widthscalefive \textwidth]{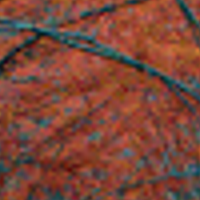} 
                    \hspace{\fsdurthree} &
                    \includegraphics[width=\widthscalefive \textwidth]{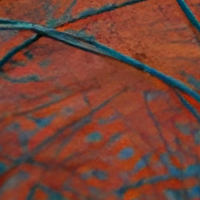} \hspace{\fsdurthree} &
                    \includegraphics[width=\widthscalefive \textwidth]{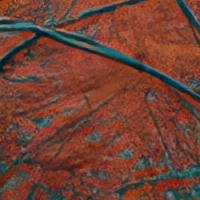} \hspace{\fsdurthree}
                    \\ 
                    LR \hspace{\fsdurthree} &
                    Real-ESRGAN~\cite{Real-ESRGAN} \hspace{\fsdurthree} &
                    \makecell{\textbf{MRDA$_{S}$-GAN}} \hspace{\fsdurthree} 
                \end{tabular}
            \end{adjustbox}
            \\
            \begin{adjustbox}{valign=t}
                \begin{tabular}{c}
                \includegraphics[height=0.55\textwidth]{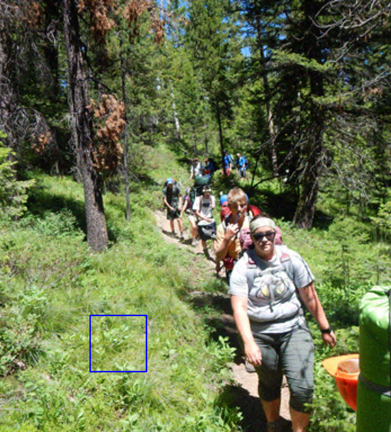}
                \end{tabular}
            \end{adjustbox}
            
            \begin{adjustbox}{valign=t}
                \begin{tabular}{ccc}
                    \includegraphics[width=\widthscalefive \textwidth]{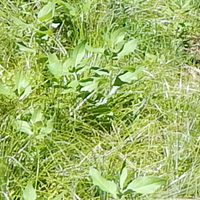} \hspace{\fsdurthree} &
                    \includegraphics[width=\widthscalefive \textwidth]{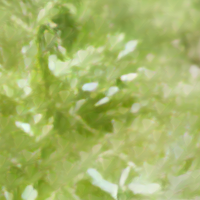} \hspace{\fsdurthree} &
                    \includegraphics[width=\widthscalefive \textwidth]{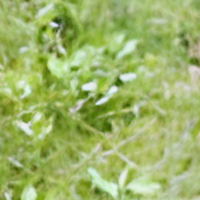} \hspace{\fsdurthree} 
                    \\
                    HR \hspace{\fsdurthree} &
                    \makecell{BSRGAN~\cite{BSRGAN}} \hspace{\fsdurthree} &
                    \makecell{MM-RealSR~\cite{MM-RealSR}} \hspace{\fsdurthree} 
                    \\
                    \includegraphics[width=\widthscalefive \textwidth]{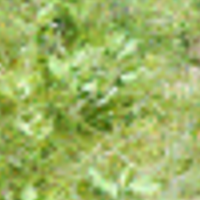} 
                    \hspace{\fsdurthree} &
                    \includegraphics[width=\widthscalefive \textwidth]{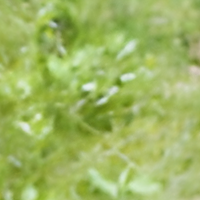} \hspace{\fsdurthree} &
                    \includegraphics[width=\widthscalefive \textwidth]{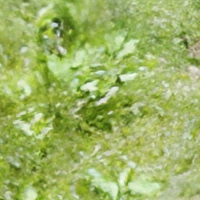} \hspace{\fsdurthree}
                    \\ 
                    LR \hspace{\fsdurthree} &
                    Real-ESRGAN~\cite{Real-ESRGAN} \hspace{\fsdurthree} &
                    \makecell{\textbf{MRDA$_{S}$-GAN}} \hspace{\fsdurthree} 
                \end{tabular}
            \end{adjustbox}

            \begin{adjustbox}{valign=t}
                \begin{tabular}{c}
                \includegraphics[height=0.55\textwidth]{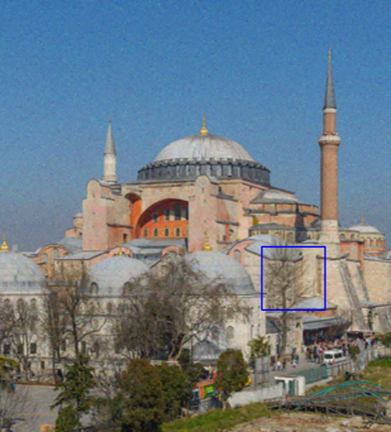}
                \end{tabular}
            \end{adjustbox}
            
            \begin{adjustbox}{valign=t}
                \begin{tabular}{ccc}
                    \includegraphics[width=\widthscalefive \textwidth]{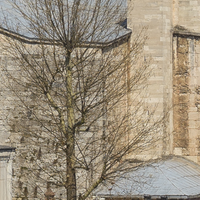} \hspace{\fsdurthree} &
                    \includegraphics[width=\widthscalefive \textwidth]{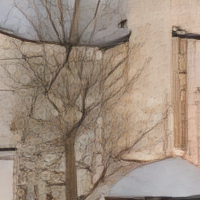} \hspace{\fsdurthree} &
                    \includegraphics[width=\widthscalefive \textwidth]{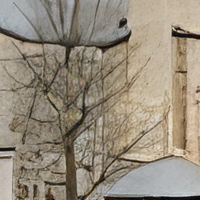} \hspace{\fsdurthree} 
                    \\
                    HR \hspace{\fsdurthree} &
                    \makecell{BSRGAN~\cite{BSRGAN}} \hspace{\fsdurthree} &
                    \makecell{MM-RealSR~\cite{MM-RealSR}} \hspace{\fsdurthree} 
                    \\
                    \includegraphics[width=\widthscalefive \textwidth]{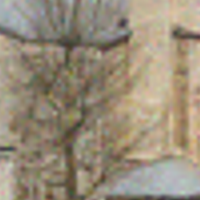} 
                    \hspace{\fsdurthree} &
                    \includegraphics[width=\widthscalefive \textwidth]{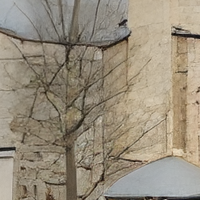} \hspace{\fsdurthree} &
                    \includegraphics[width=\widthscalefive \textwidth]{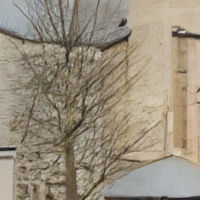} \hspace{\fsdurthree}
                    \\ 
                    LR \hspace{\fsdurthree} &
                    Real-ESRGAN~\cite{Real-ESRGAN} \hspace{\fsdurthree} &
                    \makecell{\textbf{MRDA$_{S}$-GAN}} \hspace{\fsdurthree} 
                \end{tabular}
            \end{adjustbox}
        \end{tabular}
    }
    \caption{4$\times$ visual comparison on real-world SR competition benchmarks (NTIRE2020 Track1~\cite{NTIRE2020}).}
    \label{fig:track1}
\end{figure*}

\begin{figure*}[t]
		\Huge
		\centering
		\resizebox{1\linewidth}{!}{
			\begin{tabular}{cc}
				\begin{adjustbox}{valign=t}
					\begin{tabular}{c}
		           	\includegraphics[height=1.1\textwidth]{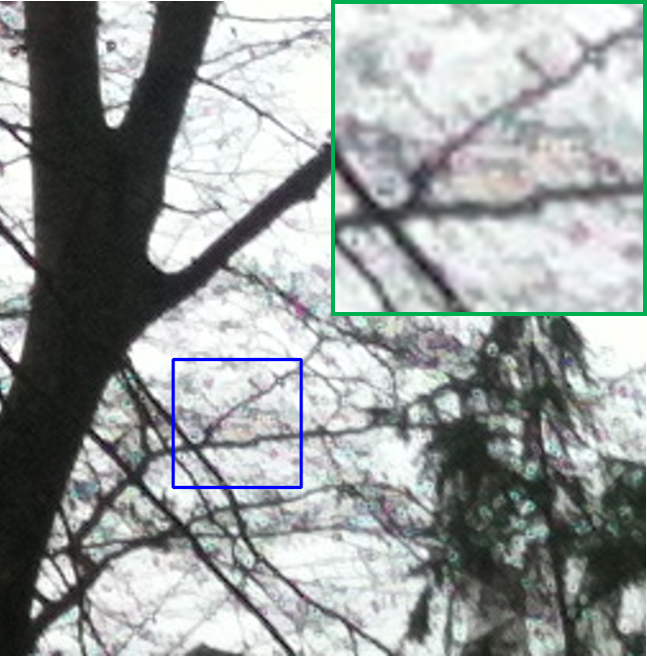}
					\end{tabular}
				\end{adjustbox}
				
				\begin{adjustbox}{valign=t}
					\begin{tabular}{cc}
						\includegraphics[width=\widthscalefivereal \textwidth]{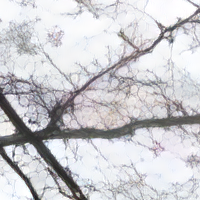} \hspace{\fsdurthree} &
						\includegraphics[width=\widthscalefivereal \textwidth]{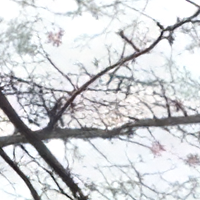} \hspace{\fsdurthree} 
						\\
						\makecell{BSRGAN~\cite{BSRGAN}} \hspace{\fsdurthree} &
						\makecell{Real-ESRGAN~\cite{Real-ESRGAN}} \hspace{\fsdurthree} 
						\\
						\includegraphics[width=\widthscalefivereal \textwidth]{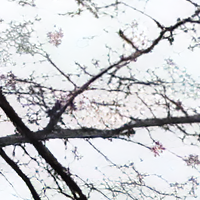} 
						\hspace{\fsdurthree} &
						\includegraphics[width=\widthscalefivereal \textwidth]{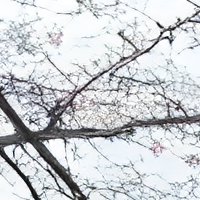} \hspace{\fsdurthree} 
						\\ 
						MM-RealSR~\cite{MM-RealSR} \hspace{\fsdurthree} &
						\makecell{\textbf{MRDA$_{S}$-GAN}} \hspace{\fsdurthree} 
					\end{tabular}
				\end{adjustbox}

				\begin{adjustbox}{valign=t}
					\begin{tabular}{c}
		           	\includegraphics[height=1.1\textwidth]{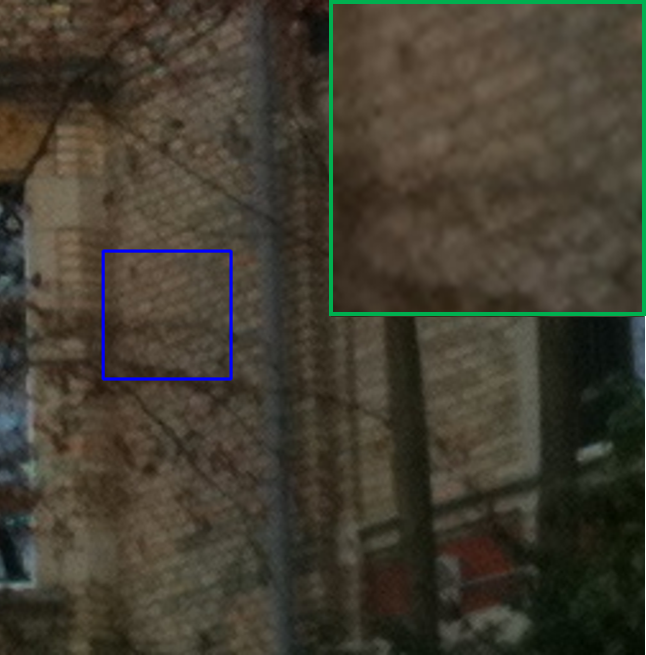}
					\end{tabular}
				\end{adjustbox}
				
				\begin{adjustbox}{valign=t}
					\begin{tabular}{cc}
						\includegraphics[width=\widthscalefivereal \textwidth]{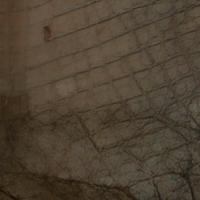} \hspace{\fsdurthree} &
						\includegraphics[width=\widthscalefivereal \textwidth]{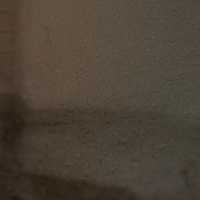} \hspace{\fsdurthree} 
						\\
						\makecell{BSRGAN~\cite{BSRGAN}} \hspace{\fsdurthree} &
						\makecell{Real-ESRGAN~\cite{Real-ESRGAN}} \hspace{\fsdurthree} 
						\\
						\includegraphics[width=\widthscalefivereal \textwidth]{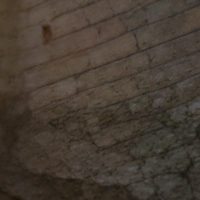} 
						\hspace{\fsdurthree} &
						\includegraphics[width=\widthscalefivereal \textwidth]{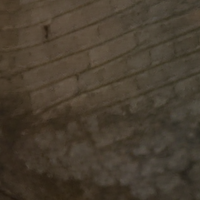} \hspace{\fsdurthree} 
						\\ 
						MM-RealSR~\cite{MM-RealSR} \hspace{\fsdurthree} &
						\makecell{\textbf{MRDA$_{S}$-GAN}} \hspace{\fsdurthree} 
					\end{tabular}
				\end{adjustbox}
                    \\
				\begin{adjustbox}{valign=t}
					\begin{tabular}{c}
		           	\includegraphics[height=1.1\textwidth]{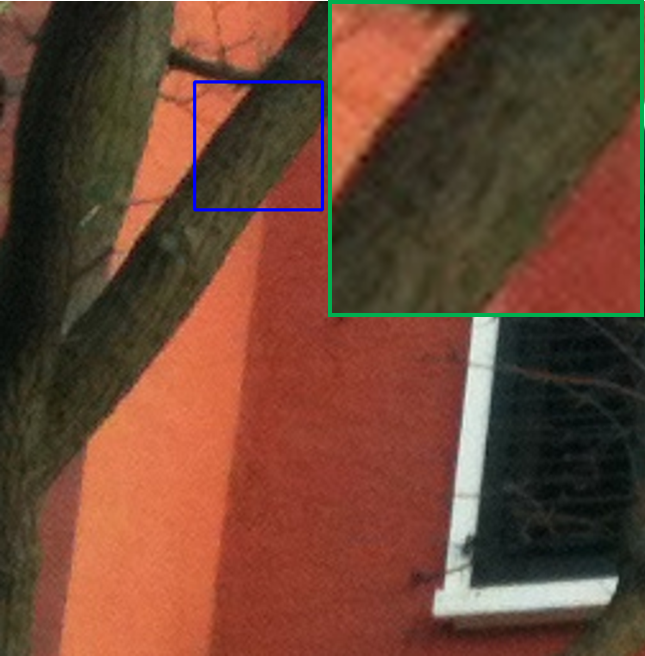}
					\end{tabular}
				\end{adjustbox}
				
				\begin{adjustbox}{valign=t}
					\begin{tabular}{cc}
						\includegraphics[width=\widthscalefivereal \textwidth]{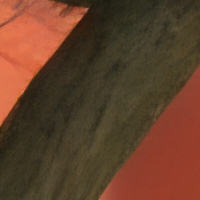} \hspace{\fsdurthree} &
						\includegraphics[width=\widthscalefivereal \textwidth]{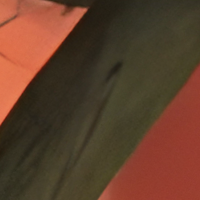} \hspace{\fsdurthree} 
						\\
						\makecell{BSRGAN~\cite{BSRGAN}} \hspace{\fsdurthree} &
						\makecell{Real-ESRGAN~\cite{Real-ESRGAN}} \hspace{\fsdurthree} 
						\\
						\includegraphics[width=\widthscalefivereal \textwidth]{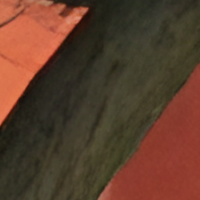} 
						\hspace{\fsdurthree} &
						\includegraphics[width=\widthscalefivereal \textwidth]{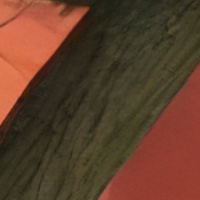} \hspace{\fsdurthree} 
						\\ 
						MM-RealSR~\cite{MM-RealSR} \hspace{\fsdurthree} &
						\makecell{\textbf{MRDA$_{S}$-GAN}} \hspace{\fsdurthree} 
					\end{tabular}
				\end{adjustbox}

				\begin{adjustbox}{valign=t}
					\begin{tabular}{c}
		           	\includegraphics[height=1.1\textwidth]{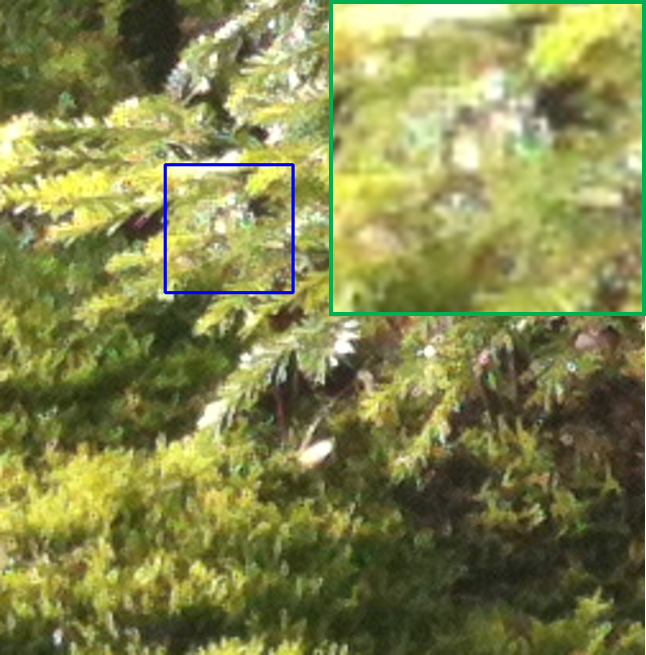}
					\end{tabular}
				\end{adjustbox}
				
				\begin{adjustbox}{valign=t}
					\begin{tabular}{cc}
						\includegraphics[width=\widthscalefivereal \textwidth]{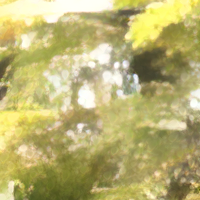} \hspace{\fsdurthree} &
						\includegraphics[width=\widthscalefivereal \textwidth]{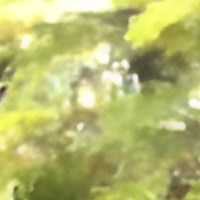} \hspace{\fsdurthree} 
						\\
						\makecell{BSRGAN~\cite{BSRGAN}} \hspace{\fsdurthree} &
						\makecell{Real-ESRGAN~\cite{Real-ESRGAN}} \hspace{\fsdurthree} 
						\\
						\includegraphics[width=\widthscalefivereal \textwidth]{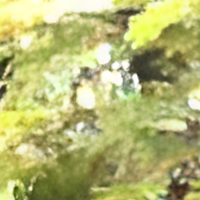} 
						\hspace{\fsdurthree} &
						\includegraphics[width=\widthscalefivereal \textwidth]{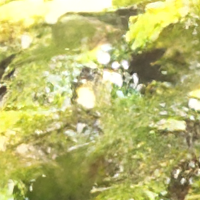} \hspace{\fsdurthree} 
						\\ 
						MM-RealSR~\cite{MM-RealSR} \hspace{\fsdurthree} &
						\makecell{\textbf{MRDA$_{S}$-GAN}} \hspace{\fsdurthree} 
					\end{tabular}
				\end{adjustbox}
			\end{tabular}
		}
		\caption{4$\times$ visual comparison on real-world SR benchmarks (NTIRE2020 Track2~\cite{NTIRE2020}, LR images of which come straight from the smartphone camera).} 
		\label{fig:track2}
	\end{figure*}

\subsection{Evaluation on Isotropic Gaussian Kernels}
\label{sec:V1}

We evaluate the performance of the proposed MRDA on classic
degradations with only isotropic Gaussian kernels. We compare the MRDA with several SR methods, including RCAN~\cite{RCAN}, SwinIR~\cite{swinir}, ZSSR~\cite{ZSSR}, IKC~\cite{IKC}, DAN~\cite{DAN}, AdaTarget~\cite{AdaTarget}, and DASR~\cite{DASR}. Note that RCAN and SwinIR  are state-of-the-art SR methods for bicubic degradation. We apply Gaussian8~\cite{IKC} kernel setting on five datasets, including Set5~\cite{Set5}, Set14~\cite{Set14}, B100~\cite{B100}, Urban100~\cite{Urban100}, and Manga109~\cite{Manga109}, to generate evaluation datasets. All methods are evaluated by PSNR on the luminance channel of the results (YCbCr space).

We compare MRDA$_{S}$ with RCAN~\cite{RCAN}, SwinIR~\cite{swinir}, ZSSR~\cite{ZSSR}, IKC~\cite{IKC}, DAN~\cite{DAN}, AdaTarget~\cite{AdaTarget}, and DASR~\cite{DASR}. The quantitative results are illustrated in Tab.~\ref{tab:iso_show}. \textbf{(1)} Although RCAN and SwinIR achieve the highest PSNR results on bicubic degradation (i.e., kernel width 0), it suffers a sharp drop in performance as the test degradations are different from the bicubic one. \textbf{(2)} Explicit degradation estimation based blind SR methods, such as IKC and DAN, adopt the iterative correction scheme to estimate degradation. Comparing IKC and MRDA$_{S}$ on PSNR, our implicit degradation estimation based MRDA$_{S}$ consumes much less time in the inference phase and surpassed IKC by 0.57~dB and 1.6~dB on Urban100 and Manga109 datasets, respectively. \textbf{(3)} Comparing both implicit degradation estimation based networks DASR and MRDA$_{S}$, our MRDA$_{S}$ surpasses DASR by 0.54~dB and 1.12~dB on Urban100~\cite{Urban100} and Manga109~\cite{Manga109} separately, which implies that the combination of meta-learning and knowledge distillation can extract more discriminative degradation than contrastive learning scheme. 

The qualitative results are shown in Fig.~\ref{fig:iso_show}, and our method has the best visual quality containing many realistic details close to respective ground-truth HR. In summary, our MRDA$_{S}$ achieves the best performance among all methods by using Meta-Learning Network (MLN) to extract implicit degradation representation (IDR), adopting knowledge distillation to learn the IDR, and employing RDAM to restore LR images.

\subsection{Evaluation on Anisotropic Gaussian Kernels and Noises}
\label{sec:V2}

In this section, we explore the performance of the proposed MRDA on classic degradations with anisotropic Gaussian kernels and noises.

We compare our method with RCAN~\cite{RCAN}, DAN~\cite{DAN}, IKC~\cite{IKC}, and DASR~\cite{DASR} on Set14~\cite{Set14}. Specifically, we adopt 9 typical blur kernels and different noise levels for evaluation. We denoise the LR images using DnCNN~\cite{DnCNN}, a state-of-the-art denoising method, for RCAN, DAN, and IKC. Besides, since IKC and Predictor are pre-trained on isotropic Gaussian kernels, we finetune them on anisotropic Gaussian kernels for fair comparison. The quantitative results are illustrated in Tab.~\ref{tab:aniso_show}.
Comparing RCAN and MRDA$_{S}$, we can see our MRDA$_{S}$ surpasses RCAN by over 1~dB as noise $\sigma=20$. Moreover, DAN and IKC are both explicit degradation estimation based methods, which can not estimate multiple degradations. We can see that our MRDA$_{S}$ surpasses IKC by over 0.9~dB under almost all degradation. Besides, comparing both implicit degradation representation based networks DASR and MRDA$_{S}$, our MRDA$_{S}$ surpasses DASR by over 0.9 dB under certain degradation.  Fig.~\ref{fig:aniso_show} shows the qualitative results. We can see that our MRDA$_{S}$ produces more visually promising results with clearer details.

\subsection{Evaluation on Real-World SR}
\label{sec:V3}

To demonstrate that our methods can be adapted to real-world degradation, we further train MRDA$_{S}$-GAN with the same high-order complex degradation process as Real-ESRGAN~\cite{Real-ESRGAN}  ( described in Sec.~\ref{sec:data_prdeal}). We introduce adversarial~\cite{Real-ESRGAN} and perceptual~\cite{perceptual} losses to train our MRDA$_{S}$-GAN (Eq.~\eqref{eq:real}). Specifically, we degrade HR images with randomly shuffled degradation, such as blur, downsampling, noise, and JPEG compression as Fig.~\ref{fig:motivation} (b) to obtain corresponding LR images for training. To validate the effectiveness of our MRDA, we conduct experiments on the dataset provided in the challenge of Real-World Super-Resolution: NTIRE2020 Track1~\cite{NTIRE2020}. Besides, we also test MRDA on RealSRSet~\cite{RealSR}, which models DLSR camera corruptions. Since NTIRE2020 track1 and RealSRSet datasets provide a paired validation
set, we can use the LPIPS~\cite{LPIPS}, PSNR, and SSIM for the evaluation.
In addition, we test our method on NTIRE2020 Track2, which was captured with smartphones. Since NTIRE2020 Track2 does not have ground-truth HR images, we mainly conduct qualitative comparisons. We compare our MRDA$_{S}$-GAN with the state-of-the-art GAN-based
SR methods, including Real-ESRGAN~\cite{Real-ESRGAN}, BSRGAN~\cite{BSRGAN}, and ESRGAN~\cite{ESRGAN}. 

The quantitative results on real-world degradations are shown in Tab.~\ref{tab:real}. We can see that our MRDA$_{S}$-GAN significantly outperforms recent real-world SR method MM-RealSR only consuming $52\%$ Mult-Adds. Furthermore, compared with Real-ESRGAN, our MRDA$_{S}$-GAN also achieves better performance consuming fewer Mult-Adds and runtime. This demonstrates the effectiveness of our MRDA on real-world SR.

The qualitative results are displayed in Figs.~\ref{fig:track1} and~\ref{fig:track2}. Specifically, we evaluate MRDA$_{S}$-GAN on real-world SR competition benchmark (NTIRE2020 Track1), and corresponding results are shown in Fig.~\ref{fig:track1}. We can see that our MRDA$_{S}$ generates more realistic details than other SOTA blind SR methods. Furthermore, we evaluate our methods on real-world images (NTIRE2020 Track2).  As shown in Fig.~\ref{fig:track2}, the restored images generated by our MRDA$_{S}$ have sharper edges and fewer artifacts.

\section{Ablation Study}
\label{sec:ablation}

\begin{figure} [t]
\setlength{\fsdttwofigBD}{-1.5mm}
\LARGE
\centering
\resizebox{1\linewidth}{!}{
	\begin{tabular}{ccc}
		\includegraphics[width=0.35\textwidth]{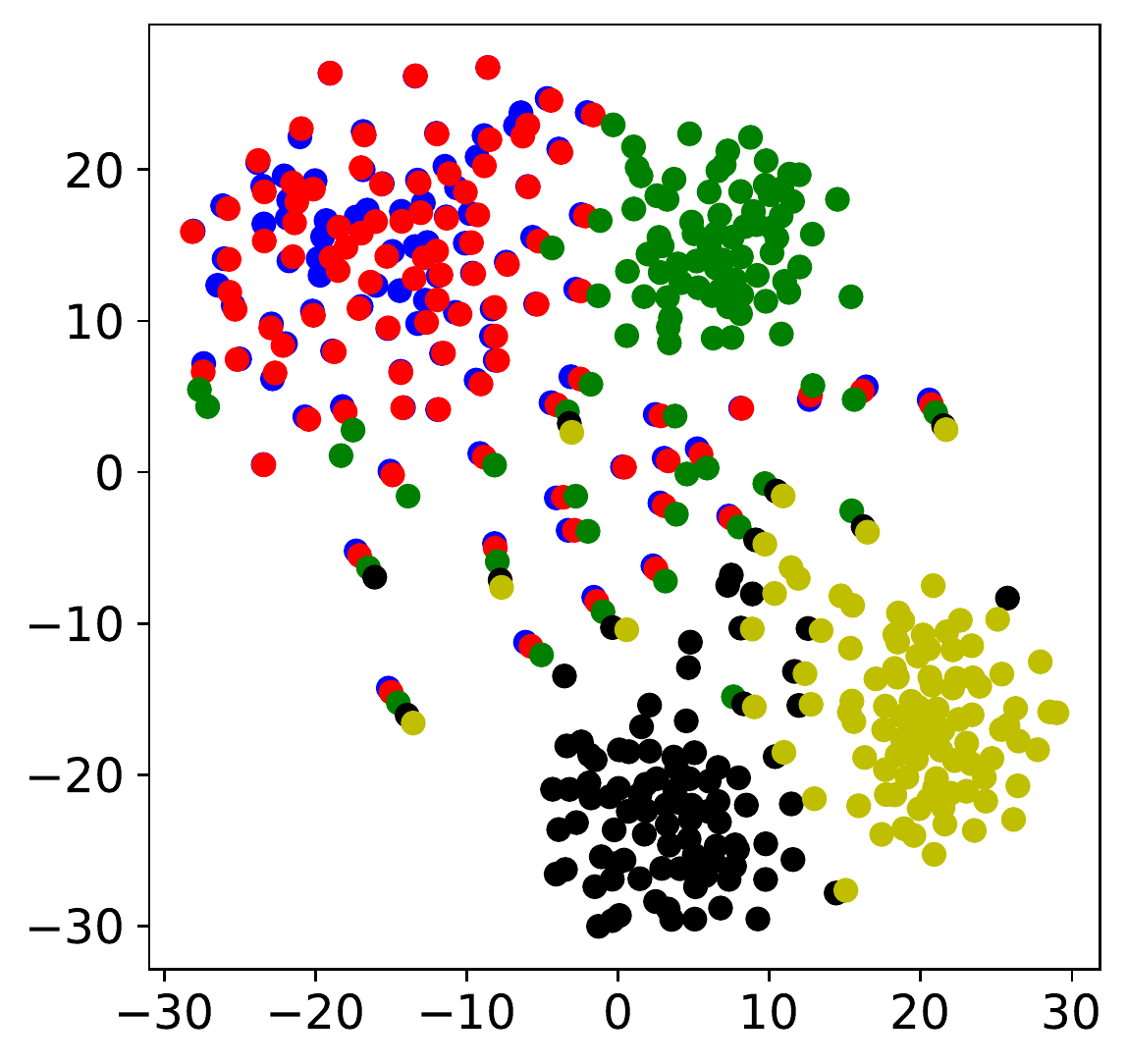} &
		\includegraphics[width=0.35\textwidth]{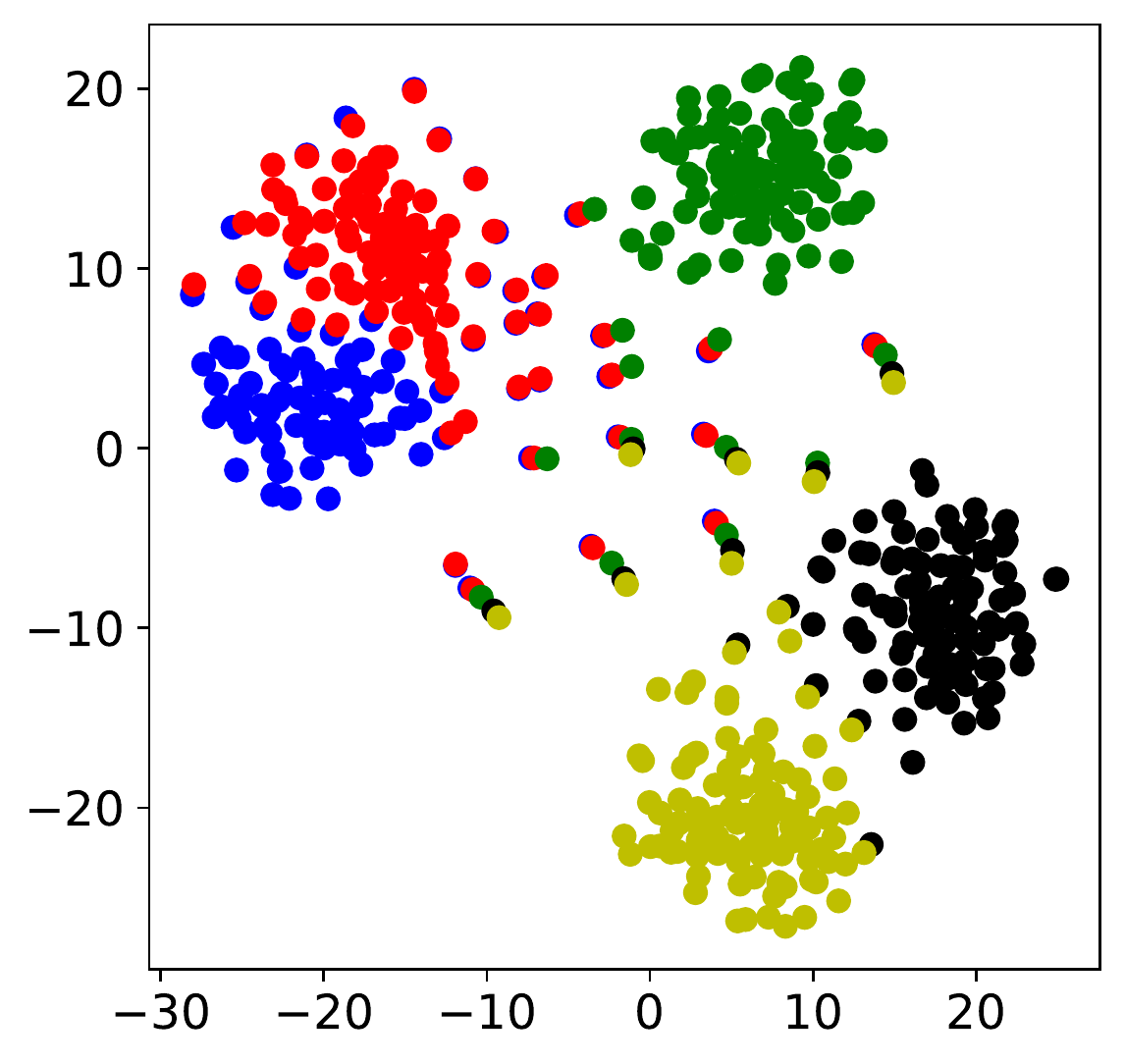} & \multirow{4}{*}[40mm]{\includegraphics[width=0.1\textwidth]{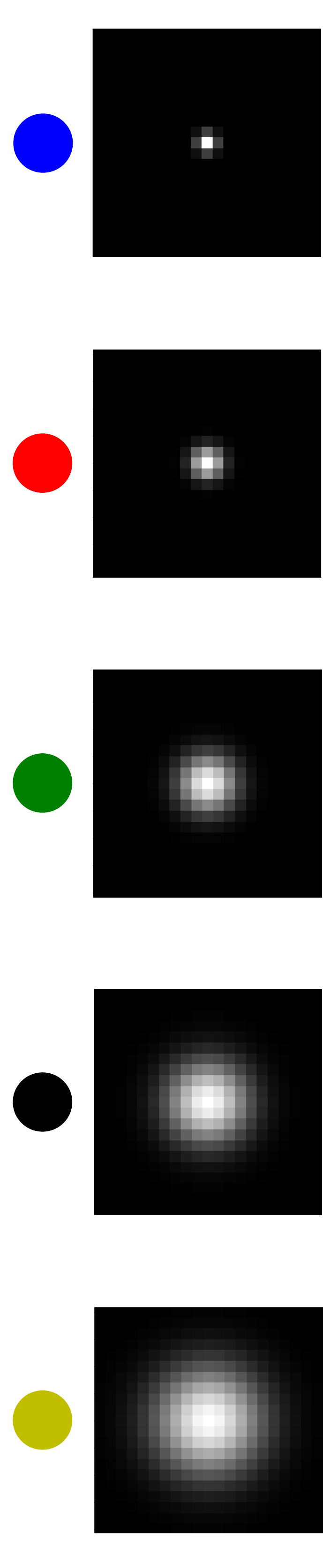}}\\
		(a) Baseline. & (b) w/ DASR. &\\
		\includegraphics[width=0.35\textwidth]{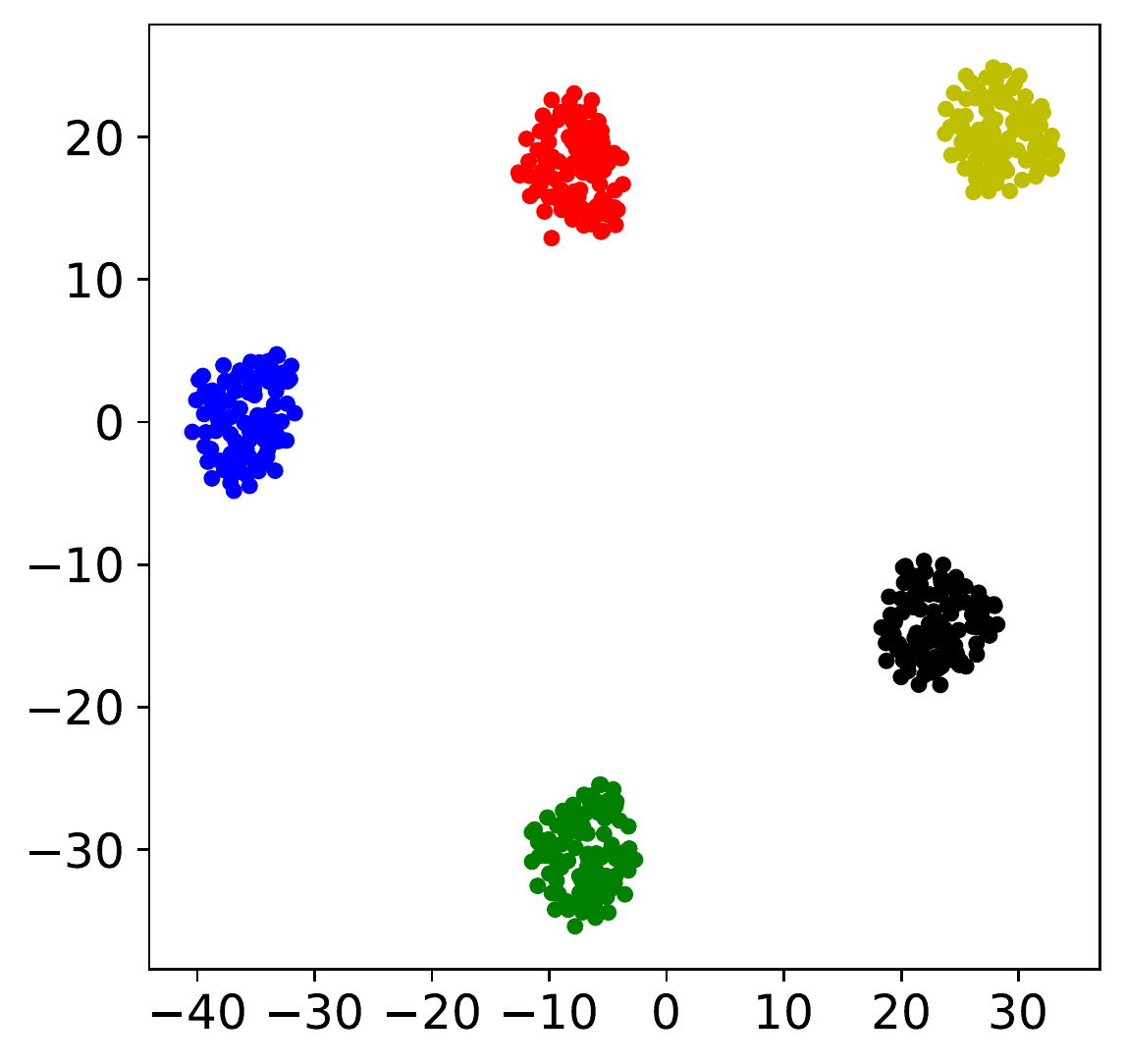} &
		\includegraphics[width=0.35\textwidth]{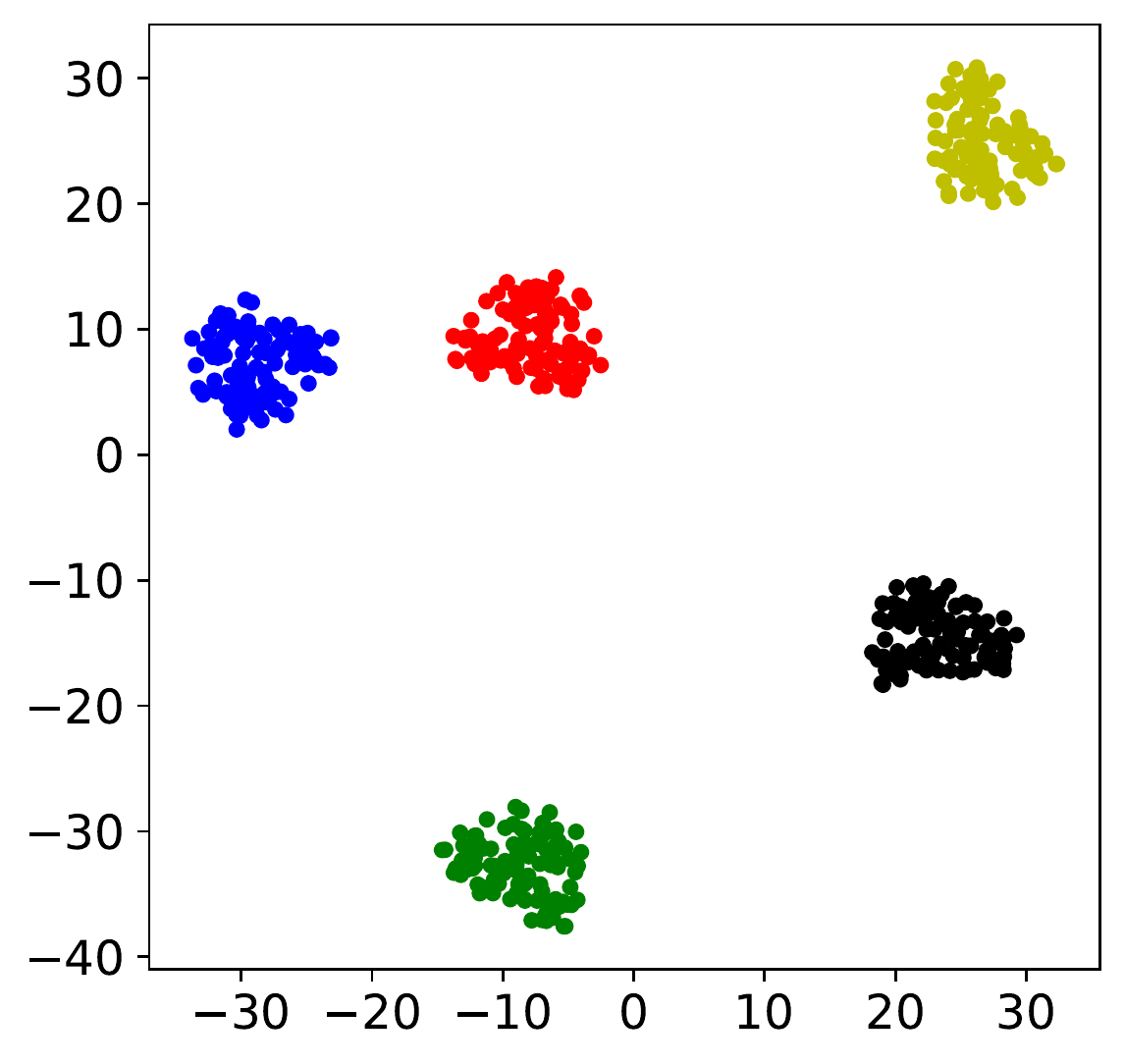} &\\
		(c) w/ DEN$_{T}$. & (d) w/ DEN$_{S}$. &
		
	\end{tabular}

}
	\caption{Visualization of implicit representations for degradations with
	different kernel widths $\sigma$. (a) shows degradation representations generated by our DEN$_{T}$ w/o degradation learning. (b) shows degradation representations generated by DASR \cite{DASR}. (c) and (d) show degradation representations generated by our DEN$_{T}$ and DEN$_{S}$. Our DEN$_{S}$ can directly extract discriminative degradation representation by learning DEN$_{T}$.}
\label{fig:iso_tsne}
\end{figure}

\begin{figure} [t]
	\setlength{\fsdttwofigBD}{-1.5mm}
	\centering
	\resizebox{1\linewidth}{!}{
		\begin{tabular}{ccc}
			\includegraphics[width=0.33\textwidth]{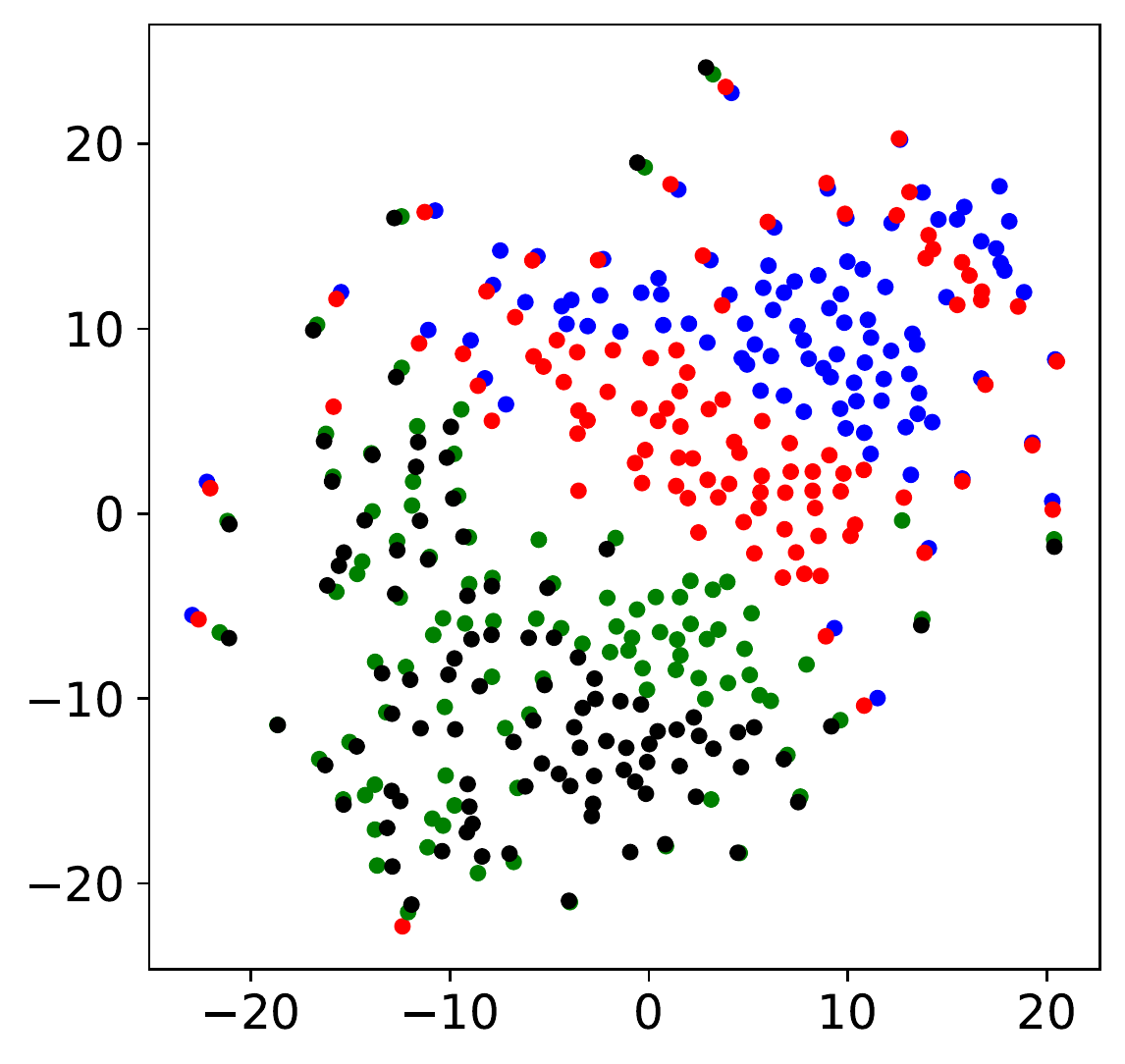} &
			\includegraphics[width=0.33\textwidth]{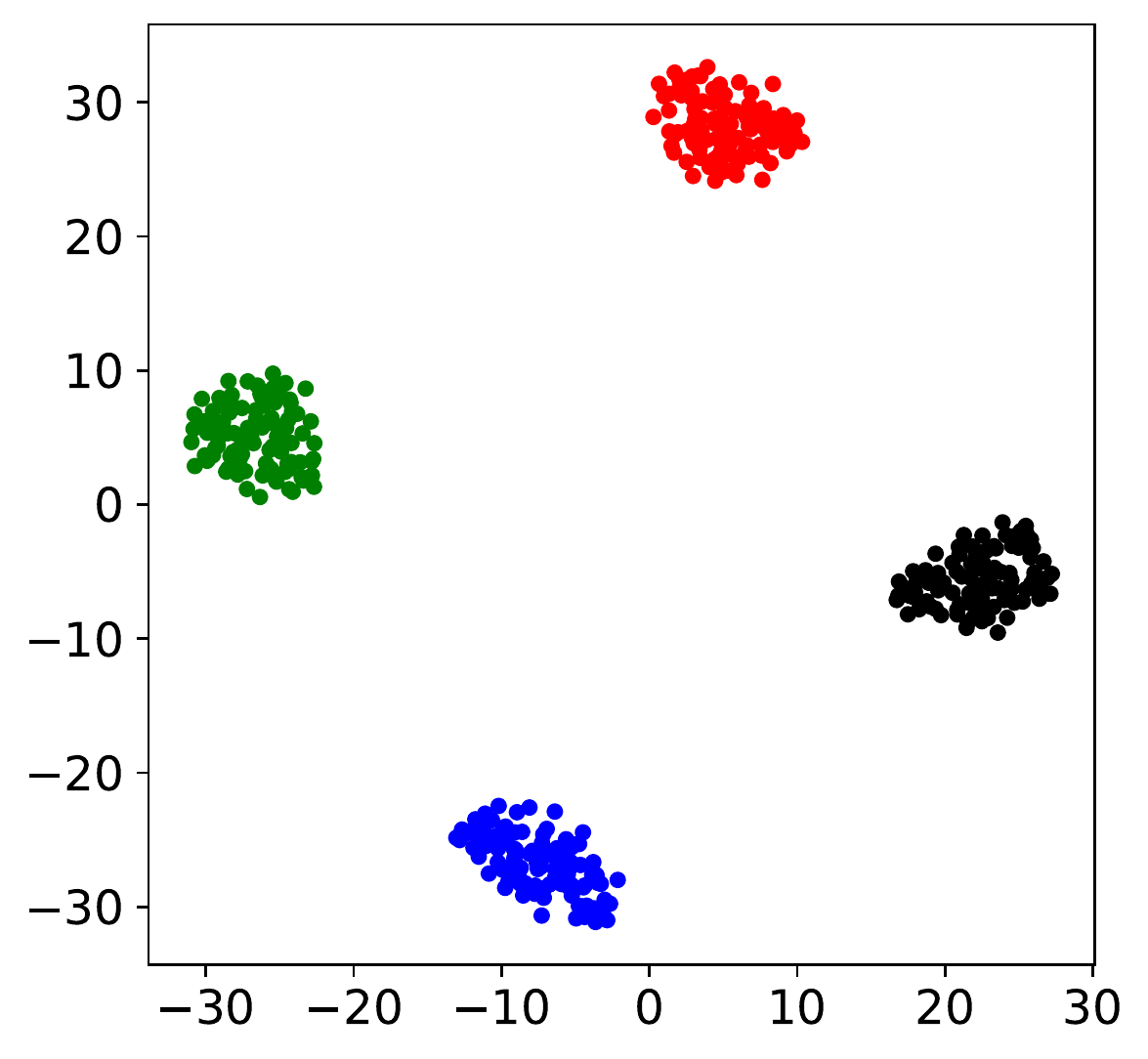} & 
 \includegraphics[width=0.065\textwidth]{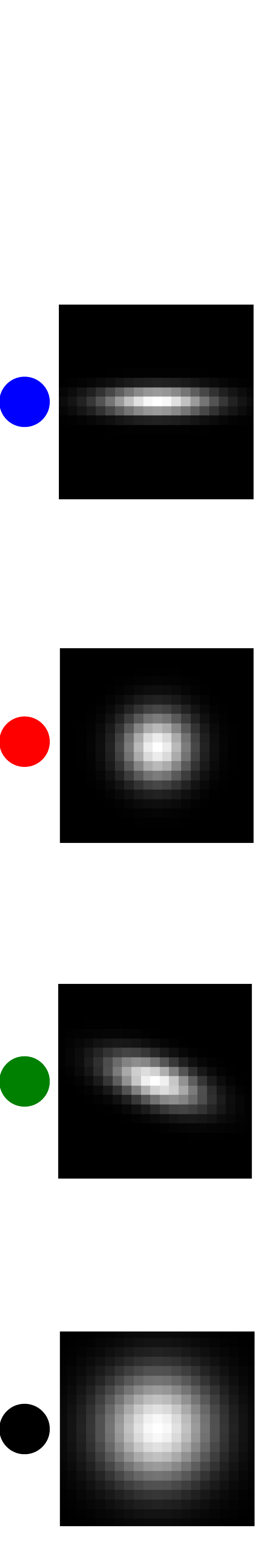}  \\
 
   			(a) w/o degradation learning (blur). & 
			(b) w/ degradation learning (blur). &  \\
   \hline
			\includegraphics[width=0.33\textwidth]{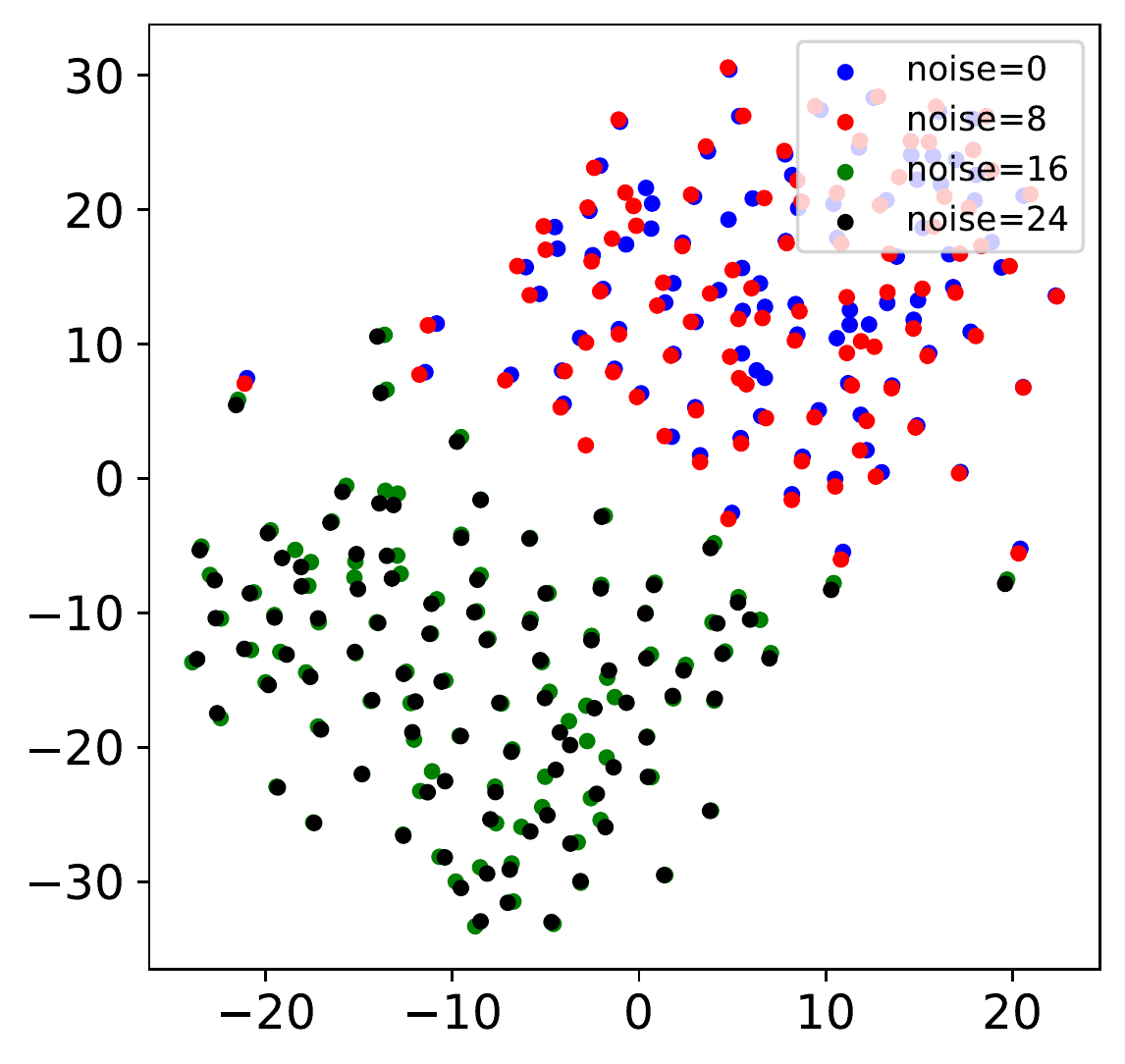} &
			\includegraphics[width=0.33\textwidth]{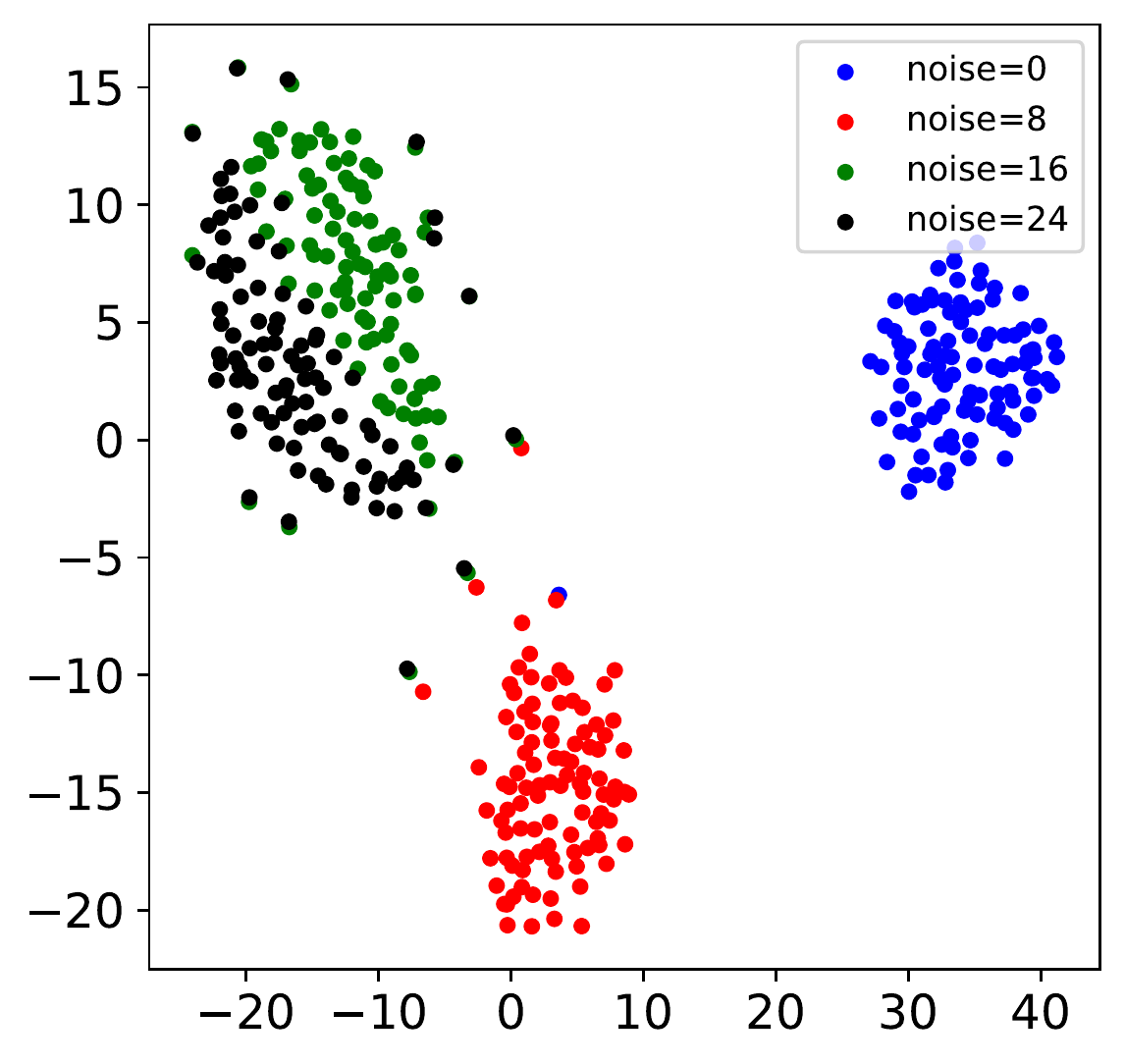} & \\
			(c) w/o degradation learning (noise). & 
			(d) w/ degradation learning (noise). &		
		\end{tabular}		
	}
	\caption{Visualization of representations for degradations with
		different anisotropic gaussian kernels (a) (c) and noise levels (b) (d).}
	\label{fig:aniso_tsne}
\end{figure}

\begin{figure*}[t]
	\setlength{\fsdttwofigBD}{-1.5mm}
	\large
	\centering
	\resizebox{0.92\linewidth}{!}{
		\begin{tabular}{cc}
			\includegraphics[width=0.46\textwidth]{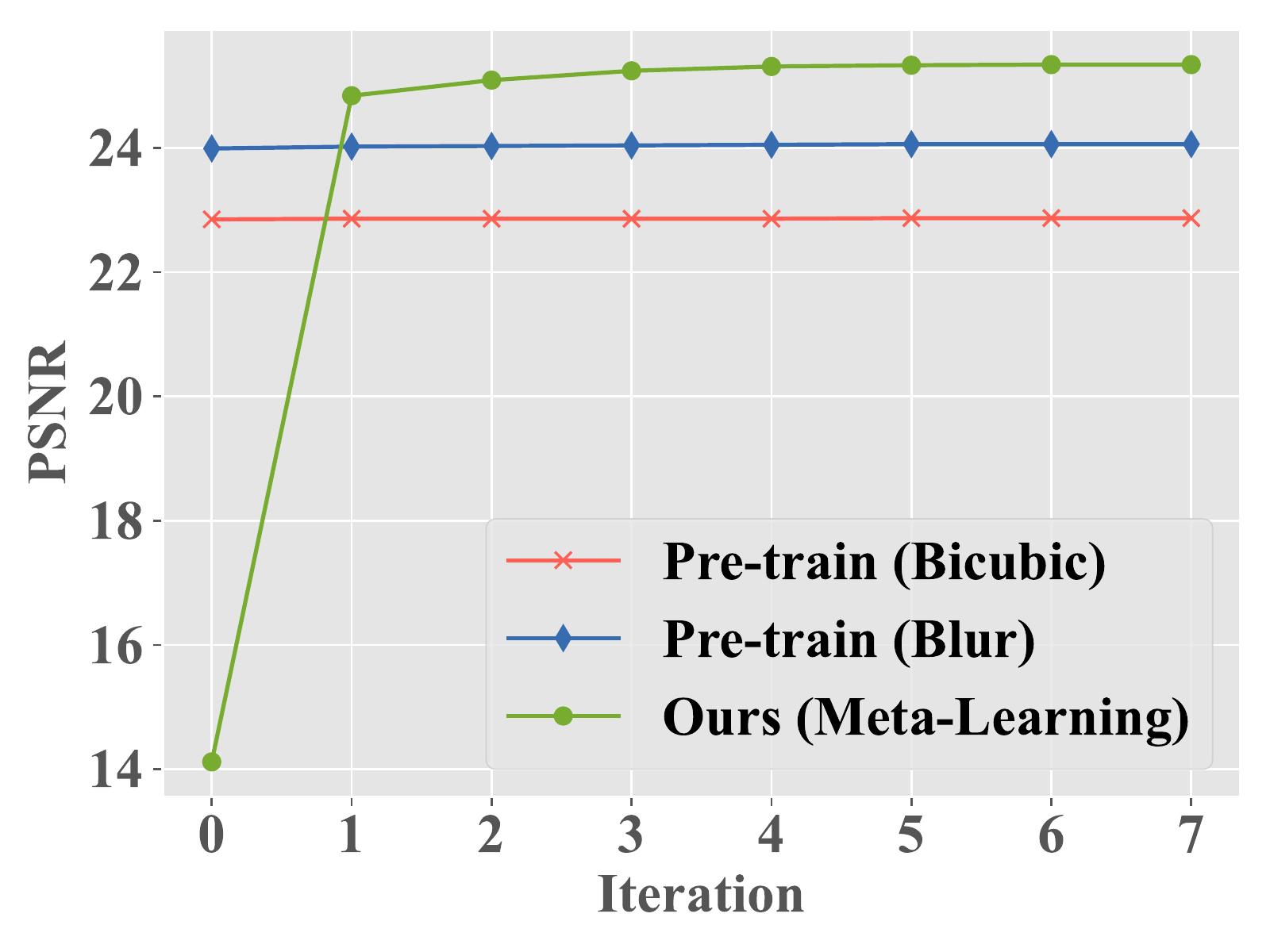} &
			\includegraphics[width=0.46\textwidth]{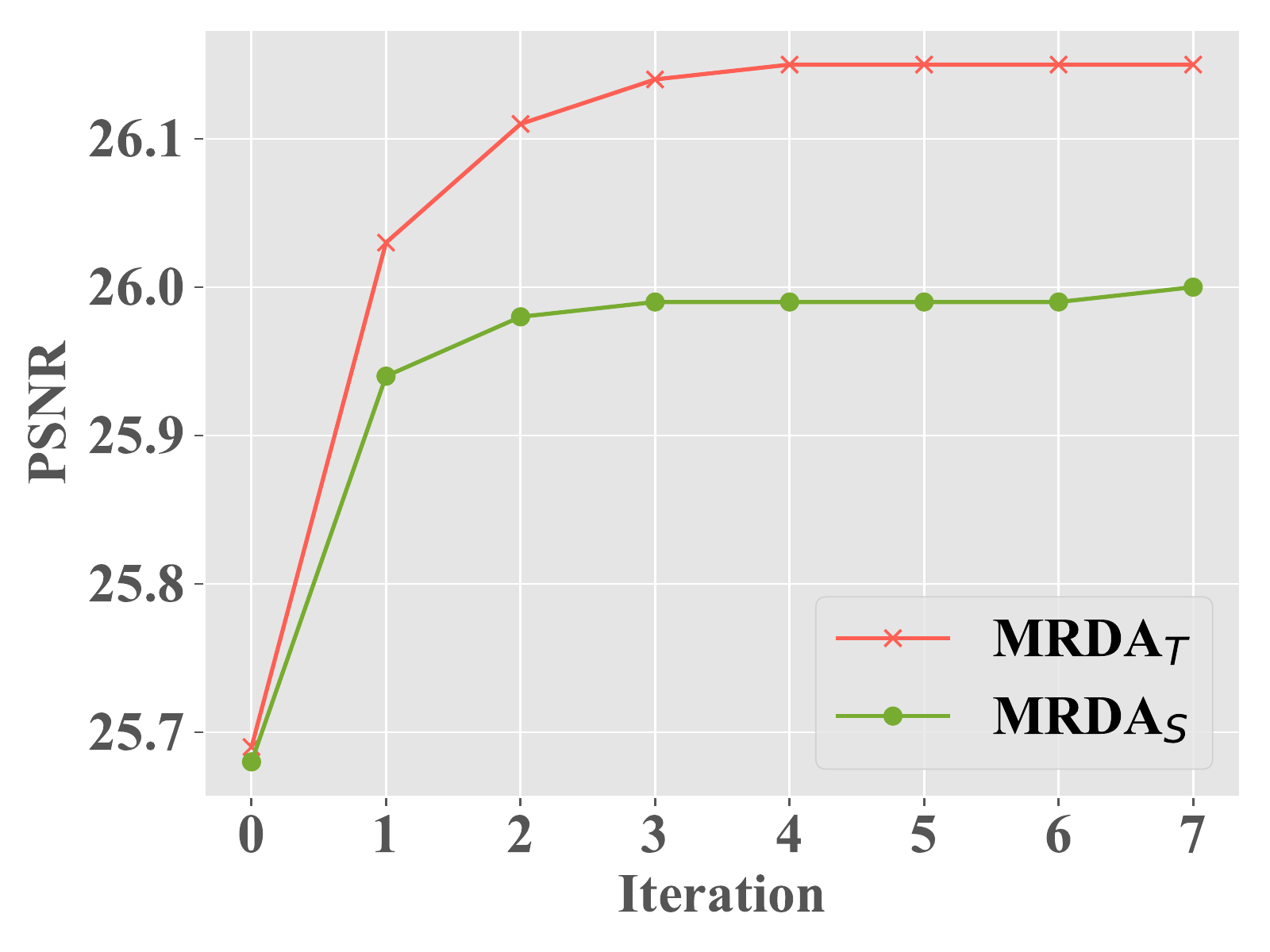} \\
			(a) The verification of the effectiveness of the meta-learning scheme. &
			(b) The relationship on the performance of MRDA and iterations.
			
		\end{tabular}
	}
	\caption{
		MLN evaluated on 4$\times$ Urban100 with Gaussian8~\cite{IKC} kernels by PSNR.}
	\label{fig:meta-learning}	
\end{figure*}

\begin{table*}[t]
	\centering
	\caption{PSNR results evaluated on Urban100 with Gaussian8~\cite{IKC} kernels for $4\times$ SR. The runtime is measured on an LR size of $180\times320$.}
	\resizebox{1\linewidth}{!}{
		\begin{tabular}{lcccccccc}
			\toprule[0.2em]
			\multicolumn{1}{c}{\multirow{2}[2]{*}{Method}} & \multicolumn{1}{c}{\multirow{2}[2]{*}{\shortstack{Meta\\Learning}}} & \multicolumn{1}{c}{\multirow{2}[2]{*}{\shortstack{Oracle \\
						Degradation}}} & \multicolumn{2}{c}{KD} & \multicolumn{1}{c}{\multirow{2}[2]{*}{RDAM }} &
			\multicolumn{1}{c}{\multirow{2}[2]{*}{\shortstack{Dynamic \\Conv} }}&
			\multicolumn{1}{c}{\multirow{2}[2]{*}{Time}} &
			\multicolumn{1}{c}{\multirow{2}[2]{*}{Urban100}} \\
			&       &       & \multicolumn{1}{c}{$\mathcal{L}_{kl}$} & \multicolumn{1}{c}{$\mathcal{L}_{abs}$} &     &  & & \\
			\midrule
			MRDA$_{T}$1 & \XSolidBrush     &   \XSolidBrush    & -     & -     & \Checkmark & \Checkmark &40.69ms  & 25.44   \\
			MRDA$_{T}$2 (Ours)&  \Checkmark    &   \XSolidBrush    & -     & -     & \Checkmark & \Checkmark &341.17ms   & 26.15  \\
			MRDA$_{T}$3 & \XSolidBrush      & \Checkmark     & -     & -     & \Checkmark   & \Checkmark &36.87ms & 26.03 \\
			\midrule
			MRDA$_{S}$1 & \XSolidBrush     &   \XSolidBrush    &  \XSolidBrush     & \Checkmark     & \Checkmark & \Checkmark & 42.61ms   & 25.83 \\
			MRDA$_{S}$2 & \XSolidBrush     &   \XSolidBrush    & \Checkmark     &   \XSolidBrush    & \Checkmark  & \Checkmark & 42.61ms  & 25.87 \\
			MRDA$_{S}$3 (Ours) & \XSolidBrush     &    \XSolidBrush   & \Checkmark     & \Checkmark     & \Checkmark  & \Checkmark & 42.61ms  & 25.90 \\
			MRDA$_{S}$4 & \XSolidBrush     &    \XSolidBrush   & \Checkmark     & \Checkmark     &    \XSolidBrush & \Checkmark & 49.74ms  & 25.86 \\
			MRDA$_{S}$5 & \XSolidBrush     &    \XSolidBrush   & \Checkmark     & \Checkmark     &    \XSolidBrush & \XSolidBrush & 41.68ms  & 24.96\\
                  MRDA$_{S}$6 (base) & \XSolidBrush     &    \XSolidBrush   & \XSolidBrush     & \XSolidBrush     &    \Checkmark & \Checkmark & 40.69ms  & 25.44\\
			\bottomrule[0.2em]
		\end{tabular}%
	}
	\label{tab:ablation}%
\end{table*}%
In this section, we further conduct ablation studies to demonstrate the effectiveness of MRDA. All ablation studies are conducted on Gaussian8~\cite{IKC} degradation setting unless otherwise stated.  

\noindent\textbf{Degradation Representation Learning.} 
In this part, we validate the effectiveness of our IDR learning. \textbf{(1)} As shown in Tab.~\ref{tab:ablation}, comparing the 1st and 2nd rows, MRDA$_{T}$2 with meta-learning based degradation representation learning performs much better than MRDA$_{T}$1 without any implicit degradation representation (IDR) learning. The experiment shows that learning discriminative IDR is essential for blind SR reconstruction. \textbf{(2)} We further visualize the degradation representation $\mathbf{D}$. Specifically, we generate LR images from BSD100 with different degradations and feed them to MRDA$_{T}$1, DASR, MRDA$_{T}$2, and MRDA$_{S}$3 to produce degradation representations for Fig.~\ref{fig:iso_tsne} (a), (b), (c), and (d) respectively, where we obtain MRDA$_{T}$1 by replacing the output of the MLN in MRDA$_{T}$2 with the LR image and changing the number of input channels in the first convolutional layer of DEN$_{T}$2 to 3. Then, these IDR are visualized using the t-SNE method~\cite{T-SNE}. As shown in Fig.~\ref{fig:iso_tsne} (a), without IDR learning, the degradation estimator fails to distinguish various degradations clearly. Comparing Fig.~\ref{fig:iso_tsne} (b) and (d), our meta-learning based IDR learning can distinguish diverse kernels more clearly than DASR. Comparing (c) and (d), the Degradation Extraction Student Network (DEN$_{S}$) can learn the degradation representation from teacher DEN$_{T}$ well. 

\noindent\textbf{Degradation Representations.} In this part, we explore the degradation representations of MRDA on anisotropic Gaussian kernels and
noises. Specifically, following the same operation in Fig. \ref{fig:iso_tsne}, we further visualize the degradation $\mathbf{D}$ extracted by DEN on BSD100~\cite{B100} using t-SNE~\cite{T-SNE}. The results are shown in Fig. \ref{fig:aniso_tsne}, where ``w/o degradation learning" uses the MRDA$_{T}$1 while ``w degradation learning" adopts the MRDA$_{S}$3 in Tab.~\ref{tab:ablation}. As shown in Fig.~\ref{fig:aniso_tsne}, our DEN$_S$ of MRDA$_{S}$3 can more clearly cluster degradations than MRDA$_{T}$1 without MLN.

\noindent\textbf{Meta-Learning Network.} To verify the effectiveness of the Meta-Learning Network (MLN), as shown in Fig.~\ref{fig:meta-learning} (a), we train the MLN with Bicubic downsampling, isotropic Gaussian kernels (blur), and our meta-learning scheme separately. The results show that the meta-learning scheme can make the MLN adapt to a specific degradation with several iterations. In addition, we explore the relationship between the performance of MRDA and the number of MLN iterations. As shown in Fig.~\ref{fig:meta-learning} (b), the performance of MRDA$_{T}$ and MRDA$_{S}$ both reaches saturation with around 5 times iterations.

\noindent\textbf{Knowledge Distillation.} The knowledge distillation (KD) is used to enforce the Degradation Extraction Student Network (DEN$_{S}$)  to learn the IDR from DEN$_{T}$. As described in Sec.~\ref{sec:distillation}, the knowledge distillation loss includes distribution loss $\mathcal{L}_{kl}$ and absolute value loss $\mathcal{L}_{abs}$. To verify the effectiveness of KD loss $\mathcal{L}_{kl}$ and $\mathcal{L}_{abs}$, we progressively add them for network training.
As shown in the 4th, 5th, and 6th rows of Tab.~\ref{tab:ablation}, the combination of $\mathcal{L}_{kl}$ and $\mathcal{L}_{abs}$ can achieve better performance than using a single loss function. Moreover,  in Tab.~\ref{tab:ablation}, MRDA$_{S}$3 surpasses MRDA$_{S}$6 by 0.46~dB. That is because MRDA$_{S}$3 further uses KD to learn and introduce implicit degradation
estimation. In addition, our MRDA$_{S}$3 exhibits comparable performance to the teacher network MRDA$_{T}$2, demonstrating that  the knowledge distillation scheme can effectively learn accurate IDR as MRDA$_{T}$2. Furthermore, unlike MRDA$_{T}$2 which requires multiple MLN iterations to extract IDR, MRDA$_{S}$3 can directly extract IDR  from LR images, resulting in significantly faster inference speeds.

\begin{figure} [t]
	\setlength{\fsdttwofigBD}{-1.5mm}
	\centering
	\resizebox{1\linewidth}{!}{
		\begin{tabular}{ccccc}
			\includegraphics[width=0.23\textwidth]{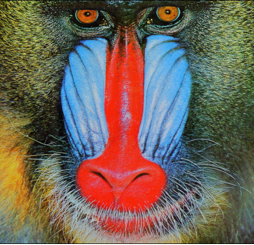} &
			\includegraphics[width=0.23\textwidth]{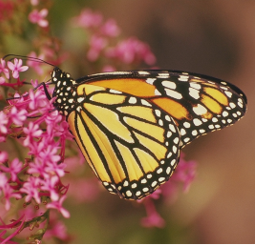} & 
			\includegraphics[width=0.23\textwidth]{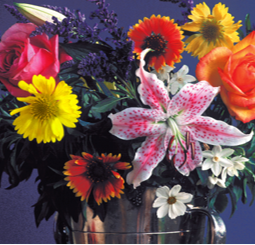} & 
			\includegraphics[width=0.23\textwidth]{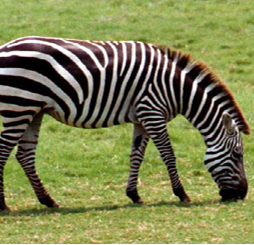} & 
			\\
			\includegraphics[width=0.23\textwidth]{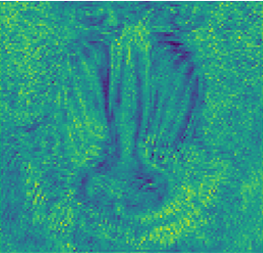} &
			\includegraphics[width=0.23\textwidth]{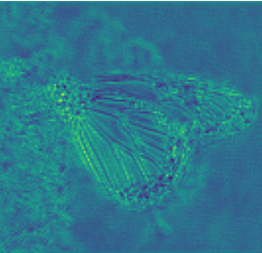} & 
			\includegraphics[width=0.23\textwidth]{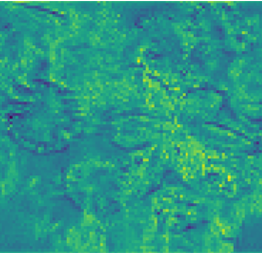} & 
			\includegraphics[width=0.23\textwidth]{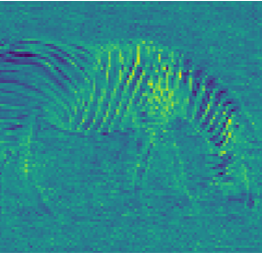} 
			
		\end{tabular}
		
	}
	\caption{ The visualization of Region Degradation Aware Modulation. The first row shows HR images, and the second row displays the modulation coefficient maps generated by RDAM.}
\label{fig:heat_map}
\end{figure}

\noindent\textbf{Region Degradation Aware Modulation.} To validate the effectiveness of Region Degradation Aware Modulation (RDAM), we replace the RDAM module of MRDA$_{S}$3  with DA module from DASR~\cite{DASR} to obtain MRDA$_{S}$4. As shown in Tab.~\ref{tab:ablation}, MRDA$_{S}$3 outperforms MRDA$_{S}$4 while consuming less runtime, which validates the effectiveness of our RDAM. Moreover, we visualize the modulation coefficient maps of Region Degradation Aware Modulation (RDAM). As shown in Fig.~\ref{fig:heat_map}, the RDAM gives larger coefficients to the region with dense textures implying that implicit degradation representation (IDR) provides more contributions to the texture restoration. In addition, we replace the dynamic convolution of MRDA$_{S}$4 with vanilla ones to obtain MRDA$_{S}$5. As shown in Tab.~\ref{tab:ablation},  MRDA$_{S}$4 significantly outperforms  MRDA$_{S}$5, which demonstrates that the dynamic convolution can effectively leverage the IDR to guide blind SR.

\noindent\textbf{Comparison with Non-Blind SR.} We further demonstrate the effectiveness of meta-learning based degradation representation and explore the upper bound performance of our MRDA by providing ground-truth degradation. Specifically, we replace the Degradation Extraction Teacher Network (DEN$_{T}$) with 5 FC layers to learn a representation directly from the true degradation (\ie, blur kernel) and train MRDA$_{T}$3 with the same setting described in Sec.~\ref{sec:data_prdeal}. As shown in the 2nd and 3rd rows of Tab.~\ref{tab:ablation}, our MRDA$_{T}$2 using implicit degradation information provided by MLN achieves even better performance than MRDA$_{T}$3 using true degradation, which further demonstrates the effectiveness of our meta-learning based degradation representation learning.

\begin{table}[t]
\centering
\caption{The verification of the effectiveness of fixed Upscaler in MLN. PSNR results achieved on Urban100 for 4$\times$ SR by MRDA$_{T}$.}
\resizebox{1\linewidth}{!}{
    \begin{tabular}{ccccc}
        \toprule[0.2em]
       Isotropic Gaussian kernel width $\sigma$ & 0   & 1.2     & 2.4   & 3.6  \\
        \midrule
        w/o fixed Upscaler in MLN    & 26.44 & 26.42 & 26.09 & 25.22  \\
        \midrule
        w/ fixed Upscaler in MLN     &    26.48   &   26.47    &   26.25    &  25.41       \\
        \bottomrule[0.2em]
    \end{tabular}%
}
\label{tab:fixed}%
\end{table}%

\noindent\textbf{Fixed parameters of Upscaler in MLN.} In the training and testing phase of the Meta-Learning Network (MLN), to keep the extracted degradation information in the same domain, we fix the parameters of Upscaler in MLN after finishing the initial pre-training on bicubic degradation (i.e., not update the parameters of Upscaler in MLN). To verify the effectiveness of this trick, we compare MRDA$_{T}$ without fixed Upscaler and with fixed Upscaler in MLN.  
As shown in Tab.~\ref{tab:fixed}, we evaluate the trick quantitatively. We can see that fixed parameters of Upscaler in MLN can bring 0.18~dB improvement.

\section{Conclusion}

In this paper, we propose a Meta-Learning based Region Degradation Aware SR Network (MRDA). Specifically, we use Meta-Learning Network (MLN) and Degradation Extraction Teacher Network (DEN$_{T}$) to extract implicit degradation representation without the supervision of ground-truth degradation. Since the MLN requires HR images, we adopt knowledge distillation to make the DEN$_{S}$ directly extract the same implicit degradation representation as DEN$_{T}$ from LR images. Moreover, we propose a Region Degradation Aware SR Network (RDAN) using spatial-wise modulation to adjust the influence of degradation representation. Our method achieves SOTA results quantitatively and qualitatively.

\bibliographystyle{IEEEtran}
\bibliography{egbib}





\vfill

\end{document}